%% file: look_listen_learn.tex
\documentclass[10pt,twocolumn,letterpaper]{article}

\usepackage{iccv}
\usepackage{times}
\usepackage{epsfig}
\usepackage{graphicx}
\usepackage{amsmath}
\usepackage{amssymb}

\pdfoutput=1
\usepackage[numbers,sort]{natbib} %
\usepackage{placeins} %
\usepackage{afterpage}
\usepackage{adjustbox} %
\usepackage{multirow}
\renewcommand{\paragraph}[1]{\smallskip\noindent{\bf{#1}}}
\usepackage[customdriver=hpdftex,pagebackref=true,breaklinks=true,letterpaper=true,colorlinks,bookmarks=false]{hyperref}

\hypersetup{pdfauthor={Relja Arandjelovic},pdftitle={Look, Listen and Learn}}

\iccvfinalcopy %

\def\iccvPaperID{191} %

\begin{document}

\title{Look, Listen and Learn}
\author{Relja Arandjelovi\'c$^\dagger$\\
{\tt\small relja@google.com}
\and
Andrew Zisserman$^{\dagger,*}$\\
{\tt\small zisserman@google.com}
\and
$^\dagger$DeepMind
\quad\quad\quad
$^*$VGG, Department of Engineering Science, University of Oxford\\
}

\maketitle

\input{floats.tex}
\input{floats_supp.tex}

\begin{abstract}
We consider the question: what can be learnt by looking at and listening to
a large number of unlabelled videos? There is a 
valuable, but so far untapped, source of information
contained in the video itself -- the correspondence between the visual
and the audio streams, and we introduce a 
novel ``Audio-Visual Correspondence'' learning task that 
makes use of this.  Training visual and audio networks from scratch,
without any additional supervision other than the raw unconstrained
videos themselves, is shown to successfully solve this task, and,
more interestingly, result in good visual and audio representations.
These features set the new state-of-the-art on two sound classification
benchmarks, and perform on par with the state-of-the-art self-supervised
approaches on ImageNet classification.
We also demonstrate that the network is able to localize objects
in both modalities, as well as perform fine-grained recognition tasks.
\end{abstract}
\vspace{-0.4cm}

\section{Introduction}

Visual and audio events tend to occur together; not always but
often: the movement of fingers and sound of the instrument when a
piano, guitar or drum is played; lips moving and speech when talking;
cars moving and engine noise when observing a street.
The visual and audio events are concurrent in these cases
because there is a common cause. 
In this paper we investigate whether we can use this simple
observation to learn about the world both visually and aurally by
simply watching and listening to videos. 

We ask the question: what can be learnt by training visual and audio
networks {\em simultaneously} to predict whether visual information
(a video frame) corresponds or not to audio information (a sound snippet)?
This is a looser requirement than that the visual and audio
events occur in sync. It only requires that there is something in
the image that correlates with something in the audio clip -- a car present
in the video frame, for instance, correlating with engine noise; or an  exterior shot with the sound of wind.

Our motivation for this work is three fold: first, as in many recent
self-supervision tasks~\cite{Dosovitskiy14,Doersch15,Agrawal15,Wang15,Zhang16,Misra16,Pathak16,Noroozi16}, it
is interesting to learn from a virtually infinite source of free
supervision (video with visual and audio modes in this case) rather
than requiring strong supervision; second, this is a possible source
of supervision that an infant could use as their visual and audio
capabilities develop; third, we want to know what can be learnt, and
how well the networks are trained, for example in the performance of
the visual and audio networks for other tasks.

Of course, we are not the first to make the observation that visual
and audio events co-occur, and to use their concurrence or correlation
as supervision for training a network. In a series of recent and
inspiring papers~\cite{Owens16,Owens16a,Aytar16,Harwath16}, the group
at MIT has investigated precisely this. However, their goal is always
to train a single network for one of the modes, for example, train a
visual network to generate sounds in~\cite{Owens16,Owens16a}; or train
an audio network to correlate with visual outputs
in~\cite{Aytar16,Harwath16}, where the visual networks are pre-trained
and fixed and act as a teacher.  In earlier, pre deep-learning,
approaches the observation was used to beautiful effect
in~\cite{Kidron05} showing ``pixels that sound'' (\eg for a guitar)
learnt using CCA. In contrast, we train both visual and audio networks and,
somewhat surprisingly, show that this is beneficial -- in that our
performance improves substantially over that of~\cite{Aytar16} when
trained on the same data.

In summary: our goal is to design a system that is able to learn both
visual and audio semantic information in a completely unsupervised
manner by simply looking at and listening to a large number of
unlabelled videos. To achieve this we introduce a novel 
\emph{Audio-Visual Correspondence (AVC)} learning task that is used
to train the two (visual and audio) networks from scratch. 
This task is described
in section~\ref{sec:approach}, together with the network architecture
and training procedure. In section~\ref{sec:results} we describe what
semantic information has been learnt, and assess the performance of
the audio and visual networks. We find, which we had not anticipated,
that this task leads to quite fine grained visual and audio
discrimination, \eg into different instruments. In terms of
quantitative performance, the audio network exceed those recently
trained for audio recognition using visual supervision, 
and the visual network has similar
performance to those trained for other, purely visual,
self-supervision tasks. Furthermore, we show, as an added benefit, that we are
able to {\em localize} the source of the audio event in the video frame (and
also localize 
the corresponding regions of the sound source) using activation
visualization.

In terms of prior work, the most closely related deep learning
approach that we know of is `SyncNet' in~\cite{Chung16a}. However,
\cite{Chung16a}~is aimed at learning to synchronize lip-regions and speech
for lip-reading, rather than the more general video and audio material
considered here for learning semantic representations.
More generally, the AVC task is a form of co-training~\cite{Blum98}, where there
are two `views' of the data, and each view provides complementary information.
In our case the two views are visual and audio (and each can determine
semantic information independently). A similar scenario arises when
the two views are visual and language (text) as
in~\cite{Frome13,Lei15,Socher13} where a common embedding is learnt. However, usually
one (or both) of the networks (for images and text) are pre-trained,  in contrast to the
approach taken here where no supervision is required and both
networks are trained from scratch.

\section{Audio-visual correspondence learning}
\label{sec:approach}
\label{sec:avc}

The core idea is to use a valuable but so far
untapped source of information contained in the video itself --
the correspondence between visual and audio streams available by
virtue of them appearing together at the same time in the same video.
By seeing and hearing many examples of a person playing a violin
and examples of a dog barking, and never, or at least very infrequently,
seeing a violin being played while hearing a dog bark and vice versa,
it should be possible to conclude what a violin and a dog look and sound
like, without ever being explicitly taught what is a violin or a dog.

\figTask

We leverage this for learning by an
\emph{audio-visual correspondence (AVC)}
task, illustrated in Figure \ref{fig:task}.
The AVC task is a simple binary classification task:
given an example video frame and a short audio clip -- decide whether
they correspond to each other or not. The corresponding (positive) pairs
are the ones that are taken at the same time from the same video,
while mismatched (negative) pairs are extracted from different videos.
The only way for a system to solve this task is if it learns
to detect various semantic concepts in both the visual and the audio domain.
Indeed, we demonstrate in Section \ref{sec:resqual} that our network
automatically learns relevant semantic concepts in both modalities.

It should be noted that the task is very difficult.
The network is made to learn visual and audio features and concepts from
scratch without ever seeing a single label.
Furthermore, the AVC task itself is quite hard when done on
completely unconstrained videos -- videos can be very noisy,
the audio source is not necessarily visible in the video
(\eg camera operator speaking, person narrating the video,
sound source out of view or occluded, \etc),
and the audio and visual content can be completely unrelated
(\eg edited videos with added music, very low volume sound,
ambient sound such as wind dominating the audio track despite
other audio events being present, \etc).
Nevertheless, the results in Section \ref{sec:results} show that our network
is able to fairly successfully solve the AVC task, and in the process
learn very good visual and audio representations.

\subsection{Network architecture}
\label{sec:network}

\figNet

To tackle the AVC task, we propose the
network structure shown in Figure \ref{fig:net}.
It has three distinct parts: the vision and the audio subnetworks
which extract visual and audio features, respectively,
and the fusion network which takes these features into account to
produce the final decision on whether the visual and audio signals correspond.
Here we describe the three parts in more detail.

\paragraph{Vision subnetwork.}
The input to the vision subnetwork is a $224 \times 224$ colour image.
We follow the VGG-network \cite{Simonyan15} design style,
with $3 \times 3$ convolutional filters,
and $2 \times 2$ max-pooling layers with stride $2$ and no padding.
The network can be segmented into four blocks of conv+conv+pool layers
such that inside each block the two conv layers have the same number of filters,
while consecutive blocks have doubling filter numbers: 64, 128, 256 and 512.
At the very end, max-pooling is performed across all spatial locations
to produce a single 512-D feature vector.
Each conv layer is followed by batch normalization \cite{Ioffe15} and
a ReLU nonlinearity.

\paragraph{Audio subnetwork.}
The input to the audio subnetwork is a 1 second sound clip converted
into a log-spectrogram (more details are provided later in this section),
which is thereafter treated as a greyscale
$257 \times 199$ image.
The architecture of the audio subnetwork is identical to the vision one
with the exception that input pixels are 1-D intensities instead of 3-D
colours and therefore the \texttt{conv1\_1} filter sizes are $3\times$ smaller
compared to the vision subnetwork. The final audio feature is also 512-D.

\paragraph{Fusion network.}
The two 512-D visual and audio features are concatenated into a 1024-D
vector which is passed through the fusion network to produce a
2-way classification output, namely, whether the vision and audio correspond
or not.
It consists of two fully connected layers, with ReLU in between them,
and the intermediate feature size of 128-D.

\subsection{Implementation details}

\paragraph{Training data sampling.}
A non-corresponding frame-audio pair is compiled by randomly sampling two
different videos and picking a random frame from one and a random 1 second
audio clip from the other.
A corresponding frame-audio pair is created by sampling a random video,
picking a random frame in that video, and then picking a random
1 second audio clip that overlaps in time with the sampled frame.
This provides additional training samples compared to simply sampling
the 1 second audio with the frame at its mid-point.
We use standard data augmentation techniques for images:
each training image is uniformly scaled such that the smallest
dimension is equal to 256, followed by random cropping into $224 \times 224$,
random horizontal flipping, and brightness and saturation jittering.
Audio is only augmented by changing the volume up to 10\% randomly but
consistently across the sample.

\paragraph{Log-spectrogram computation.}
The 1 second audio is resampled to 48 kHz, and a spectrogram is computed
with window length of 0.01 seconds and a half-window overlap;
this produces 199 windows with 257 frequency bands.
The response map is passed through a logarithm
before feeding it into the
audio subnetwork.

\paragraph{Training procedure.}
We use the Adam optimizer \cite{Kingma15}, weight decay $10^{-5}$,
and perform a grid search on the learning rate, although $10^{-4}$
usually works well.
The network was trained on 16 GPUs in parallel with synchronous training
implemented in TensorFlow, where each worker processed a 16-element batch,
thus making the effective batch size of 256.
For a training set of 400k 10 second videos, the network is trained
for two days, during which it has seen 60M frame-audio pairs.

\section{Results and discussion}
\label{sec:results}

Our ``look, listen and learn'' network ($L^3$-Net) approach is evaluated and examined
in multiple ways.
First, the performance of the network on the audio-visual correspondence
task itself is investigated, and compared to supervised baselines.
Second, the quality of the learnt visual and audio features is tested
in a transfer learning setting, on visual and audio classification tasks.
Finally, we perform a qualitative analysis of what the network has learnt.
We start by introducing the datasets used for training.

\subsection{Datasets}
\label{sec:datasets}

Two video datasets are used for training the networks:
Flickr-SoundNet and Kinetics-Sounds.

\paragraph{Flickr-SoundNet \cite{Aytar16}.}
This is a large unlabelled dataset of completely unconstrained videos
from Flickr, compiled by searching for popular tags, but no tags
or any sort of additional information apart from the videos themselves
are used.
It contains over 2 million videos but for practical reasons we
use a random subset of 500k videos
(400k training, 50k validation and 50k test)
and only use the first 10 seconds of each video.
This is the dataset that is used for training the $L^3$-Net
for the transfer learning experiments in
Sections \ref{sec:resaudio} and \ref{sec:resvisual}.

\paragraph{Kinetics-Sounds.}
While our goal is to learn from completely unconstrained videos,
having a labelled dataset is useful for quantitative evaluation.
For this purpose we took a subset (much smaller than Flickr-SoundNet) of
the Kinetics dataset \cite{Kay17}, which contains YouTube videos
manually annotated for human actions using Mechanical Turk, and cropped
to 10 seconds around the action.
The subset contains 19k 10 second video clips
(15k training, 1.9k validation, 1.9k test)
formed by filtering the Kinetics dataset for 34 human action
classes, which have been chosen to be potentially manifested
visually and aurally, such as
playing various instruments (guitar, violin, xylophone, \etc),
using tools (lawn mowing, shovelling snow, \etc),
as well as performing miscellaneous actions
(tap dancing, bowling, laughing, singing, blowing nose, \etc);
the full list is given in appendix \ref{sec:kineticssounds}.
Although this dataset is fairly clean by construction, it still contains
considerable noise, \eg the bowling action is often accompanied by loud music
at the bowling alley, human voices (camera operators or video narrations)
often masks the sound of interest, and many videos contain sound tracks
that are completely unrelated to the visual content
(\eg music montage for a snow shovelling video).

\subsection{Audio-visual correspondence}

First we evaluate the performance of our method on the task it was trained
to solve -- deciding whether a frame and a 1 second audio clip
correspond (Section \ref{sec:avc}).
For the Kinetics-Sounds dataset which contains labelled videos,
we also evaluate two supervised baselines in order to gauge how well
the AVC training compares to supervised training.

\paragraph{Supervised baselines.}
For both baselines we first train vision and audio networks independently
on the action classification task, and then combine them in two different ways.
The vision network has an identical feature extraction trunk as our vision
subnetwork (Section \ref{sec:network}),
on top of which two fully connected layers
are attached (sizes: $512 \times 128$ and $128 \times 34$) to perform
classification into the 34 Kinetics-Sounds classes.
The audio classification network is constructed analogously.
The {\bf direct combination} baseline computes the audio-video correspondence
score as the similarity of class score distributions of the two networks,
computed as the scalar product between the 34-D network softmax outputs,
and decides that audio and video are in correspondence if the score is larger
than a threshold.
The motivation behind this baseline is that if the vision network believes the
frame contains a dog while the audio network is confident it hears a violin,
then the (frame, audio) pair is unlikely to be in correspondence.
The {\bf supervised pretraining} baseline takes the feature extraction
trunks from the two trained networks, assembles them into our network
architecture by concatenating the features and adding two fully connected
layers (Section \ref{sec:network}). The weights of the feature extractors
are frozen and the fully connected layers are trained on the AVC task
in the same manner as our network.
This is the strongest baseline as it directly corresponds to our method,
but with features learnt in a fully supervised manner.

\tabAvc
\paragraph{Results and discussion.}
Table \ref{tab:avc} shows the results on the AVC task.
The $L^3$-Net achieves 74\% and 78\% on the two datasets, where chance is 50\%.
It should be noted that the task itself is quite hard due to the
unconstrained nature of the videos (Section \ref{sec:avc}),
as well as due to the very local input data which lacks context --
even humans find it hard to judge whether an isolated frame and
an isolated single second of audio correspond;
informal human tests indicated that humans are only a few percent better than the $L^3$-Net.
Furthermore, the supervised baselines do not beat the $L^3$-Net
as ``supervised pretraining'' performs on par with it, while
``supervised direct combination'' works significantly worse
as, unlike ``supervised pretraining'', it has not been trained for the AVC task.

\subsection{Audio features}
\label{sec:resaudio}

In this section we evaluate the power of the audio representation that emerges
from the $L^3$-Net approach.
Namely, the $L^3$-Net audio subnetwork trained on Flickr-SoundNet
is used to extract features from
1 second audio clips, and the effectiveness of these features is evaluated
on two standard sound classification benchmarks: ESC-50 and DCASE.

\paragraph{Environmental sound classification (ESC-50) \cite{Piczak15}.}
This dataset contains 2000 audio clips, 5 seconds each, equally balanced
between 50 classes. These include animal sounds, natural soundscapes,
human non-speech sounds, interior/domestic sounds, and exterior/urban noises.
The data is split into 5 predefined folds and performance is measured in terms
of mean accuracy over 5 leave-one-fold-out evaluations.

\paragraph{Detection and classification of acoustic scenes and events (DCASE) \cite{Stowell15}.}
We consider the scene classification task of the challenge which contains
10 classes
(bus, busy street, office, open air market, park, quiet street,
restaurant, supermarket, tube, tube station),
with 10 training and 100 test clips per class, where each clip is 30 seconds long.

\tabSound
\tabImageNet

\paragraph{Experimental procedure.}
To enable a fair direct comparison with the current state-of-the-art,
Aytar \etal \cite{Aytar16},
we follow the same experimental setup.
Multiple overlapping subclips are extracted from each recording and described
using our features. For 5 second recordings from ESC-50 we extract 10
equally spaced 1 second subclips, while for the 6 times longer DCASE recordings,
60 subclips are extracted per clip.
The audio features are obtained by max-pooling the last convolutional layer
of the audio subnetwork (\texttt{conv4\_2}), before the ReLU,
into a $4 \times 3 \times 512 = 6144$ dimensional representation
(the \texttt{conv4\_2} outputs are originally $16 \times 12 \times 512$).
The features are preprocessed using z-score normalization, \ie shifted and
scaled to have a zero mean and unit variance.
A multi-class one-vs-all linear SVM is trained, and at test time the class scores
for a recording are computed as the mean over the class scores for its subclips.

\paragraph{Results and discussion.}
Table \ref{tab:sound} shows
the results on ESC-50 and DCASE. %
On both benchmarks we convincingly beat the previous
state-of-the-art, SoundNet \cite{Aytar16}, by 5.1\% and 5\% absolute.
For ESC-50 we reduce the gap between the previous best result and the human
performance by 72\% while for DCASE we reduce the error by 42\%.
The results are especially impressive as SoundNet uses
two vision networks trained in a fully supervised manner on
ImageNet and Places2 as teachers for the audio network, while we learn
both the vision and the audio networks without any supervision whatsoever.
Note that we train our networks with a random subset of the SoundNet videos
for efficiency purposes, so it is possible that further gains can be achieved
by using all the available training data.

\subsection{Visual features}
\label{sec:resvisual}

\label{sec:imagenet}

In this section we evaluate the power of the visual representation that emerges
from the $L^3$-Net approach.
Namely, the $L^3$-Net vision subnetwork trained on Flickr-SoundNet
is used to extract features from
images, and the effectiveness of these features is evaluated
on the ImageNet large scale visual recognition challenge 2012 \cite{Russakovsky15}.

\paragraph{Experimental procedure.}
We follow the experimental setup of Zhang \etal \cite{Zhang16}
where features are extracted from $256 \times 256$ images
and used to
perform linear classification on ImageNet.
As in \cite{Zhang16},
we take \texttt{conv4\_2} features after ReLU and perform max-pooling
with equal kernel and stride sizes until feature dimensionality is below 10k;
in our case this results in $4 \times 4 \times 512 = 8192$-D features.
A single fully connected layer is added to perform linear classification
into the 1000 ImageNet classes. All the weights are frozen to their $L^3$-Net-trained
values, apart from the final classification layer which is trained with
cross-entropy loss on the ImageNet training set.
The training procedure
(data augmentation, learning rate schedule, label smoothing)
is identical to \cite{Szegedy16}, the only differences being that we use
the Adam optimizer instead of RMSprop,
and a $256 \times 256$ input image instead of $299 \times 299$
as it fits our architecture better and to be consistent with \cite{Zhang16}.

\figUnitsV
\figUnitsVmap
\afterpage{\FloatBarrier}

\paragraph{Results and discussion.}
Classification accuracy on the ImageNet validation set is shown in
Table \ref{tab:imagenet} and contrasted with other unsupervised and
self-supervised methods.
We also test the performance of random features,
\ie our $L^3$-Net architecture without AVC training but with a trained classification
layer.

Our $L^3$-Net-trained features achieve 32.3\% accuracy which is on par with other
state-of-the-art self-supervised methods of \cite{Doersch15,Zhang16,Donahue17,Noroozi16},
while convincingly beating random initialization,
data-dependent initialization \cite{Krahenbuhl15}, and
Context Encoders \cite{Pathak16}.
It should be noted that these methods use the AlexNet \cite{Krizhevsky12} architecture
which is different to ours,
so the results are not fully comparable.
On the one hand, our architecture when trained from scratch in its entirety
achieves a higher performance (59.2\% vs AlexNet's 51.0\%).
On the other hand, it is deeper which makes it harder to train
as can be seen from the fact that our random features perform worse
than theirs (12.9\% vs AlexNet's 18.3\%), and that all competing methods hit peak
performance when they use earlier layers
(\eg \cite{Donahue17} drops from 31.0\% to 27.1\%
when going from \texttt{conv3} to \texttt{pool5}).
In fact, when measuring the improvement achieved due to AVC or self-supervised
training versus
the performance of the network with random initialization,
our AVC training beats all competitors.

Another important fact to consider is that \emph{all}
competing methods actually use ImageNet images when
training. Although they do not make use of the labels,
the underlying image statistics are the same:
objects are fairly central in the image, and the networks have seen,
for example, abundant images of 120 breads of dogs and thus potentially
learnt their distinguishing features.
In contrast, we use a completely separate source of training data in the
form of frames from Flickr videos -- here the objects are in general not
centred, it is likely that the network has never seen a ``Tibetan terrier''
nor the majority of other fine-grained categories.
Furthermore, video frames have vastly different low-level statistics to
still images, with strong artefacts such as motion blur.
With these factors hampering our network, it is impressive that our
visual features $L^3$-Net-trained on Flickr videos
perform on par with self-supervised state-of-the-art trained on ImageNet.

\subsection{Qualitative analysis}
\label{sec:resqual}

In this section we analyse what is it that the network has learnt.
We visualize the results on the test set of the Kinetics-Sounds and Flickr-SoundNet datasets,
so the network has not seen the videos during training.

\subsubsection{Vision features}
To probe what the vision subnetwork has learnt, we pick a particular `unit'
in \texttt{pool4} (\ie a component of the 512 dimensional \texttt{pool4} vector) and rank the test images by its magnitude.
Figure \ref{fig:unitsV} shows the images from Kinetics-Sounds that activate particular
units in \texttt{pool4} the most (\ie are ranked highest by its magnitude). As can be
seen, the vision subnetwork has automatically learnt, without any
explicit supervision, to recognize semantic entities such as guitars,
accordions, keyboards, clarinets, bowling alleys, lawns or lawnmowers,
\etc.  Furthermore, it has learnt finer-grained categories as well as
it is able to distinguish between acoustic and bass guitars
(``fingerpicking'' is mostly associated with acoustic guitars).

Figure \ref{fig:unitsVmap} shows heatmaps for the Kinetics-Sounds images in 
Figure~\ref{fig:unitsV}, obtained by simply displaying the spatial
activations of the corresponding vision unit (\ie if the $k$ component of
\texttt{pool4} is chosen, then the $k$ channel of \texttt{conv4\_2} is displayed -- since the $k$ component is just the spatial max over this channel 
(after ReLU)).
 Objects are successfully
detected despite significant clutter and occlusions.  It is
interesting to observe the type of cues that the network decides to
use, \eg the ``playing clarinet'' unit, instead of trying to detect
the entire clarinet, seems to mostly activate on the interface between
the player's face and the clarinet.

Figures \ref{fig:unitsVSoundNetA} and \ref{fig:unitsVSoundNetB} show
visual concepts learnt by the $L^3$-Net on the Flickr-SoundNet dataset.
It can be seen that the network learns to recognize many scene categories
(Figure \ref{fig:unitsVSoundNetA}),
such as outdoors, concert, water, sky, crowd, text, railway, \etc.
These are useful for the AVC task as, for example, crowds indicate a large
event that is associated with a distinctive sound as well
(\eg a football game), text indicates narration, and outdoors scenes are
likely to be accompanied with wind sounds.
It should be noted that though at first sight some categories seem trivially
detectable, it is not the case; for example, ``sky'' detector is not equivalent to
the ``blueness'' detector as it only fires on ``sky'' and not on ``water'',
and furthermore there are separate units sensitive to ``water surface''
and to ``underwater'' scenes.
The network also learns to detect people as user uploaded content is
substantially people-oriented -- Figure \ref{fig:unitsVSoundNetB} shows
the network has learnt to distinguish between babies, adults and crowds.

\figUnitsVSoundNetA

\subsubsection{Audio features}
Figure \ref{fig:unitsA} shows what particular audio units
are sensitive to in the Kinetics-Sounds dataset.
For visualization purposes, instead of showing
the sound form, we display the video frame that corresponds to the sound.
It can be seen that the audio subnetwork, again without any supervision,
manages to learn various semantic entities, as well as perform fine-grained
classification (``fingerpicking'' vs ``playing bass guitar'').
Note that some units are naturally confused -- the ``tap dancing'' unit also
responds to ``pen tapping'', while the ``saxophone'' unit is sometimes
confused with a ``trombone''. These are reasonable mistakes, especially
when taking into account that the sound input is only one second in length.
The audio concepts learnt on the Flickr-SoundNet dataset (Figure \ref{fig:unitsASoundNetA}) follow
the same pattern as the visual ones -- the network learns to distinguish
various scene categories such as water, underwater, outdoors and windy scenes,
as well as human-related concepts like baby and human voices, crowds, \etc.

Figure \ref{fig:unitsAmap} shows spectrograms and their semantic heatmaps,
illustrating that our $L^3$-Net learns to detect audio events.
For example, it shows clear preference for low frequencies when detecting
bass guitars, attention to wide frequency range when detecting lawnmowers,
and temporal `steps' when detecting fingerpicking and tap dancing.

\figUnitsVSoundNetB
\figUnitsA
\figUnitsASoundNetA

\afterpage{\FloatBarrier}

\subsubsection{Versus random features}
Could the results in Figures
\ref{fig:unitsV},
\ref{fig:unitsVmap},
\ref{fig:unitsVSoundNetA},
\ref{fig:unitsVSoundNetB},
\ref{fig:unitsA}
\ref{fig:unitsASoundNetA},
and
\ref{fig:unitsAmap}
simply be obtained by chance due to examining a large number of units,
as colourfully illustrated by the dead salmon experiment
\cite{Bennett09}?  It is unlikely as there are only 512 units in
\texttt{pool4} to choose from, and many of those were found to be
highly correlated with a semantic concept. Nevertheless, we repeated
the same experiment with a random network (\ie a network that has not
been trained), and have failed to find such correlation.  In more
detail, we examined how many out of the action classes in
Kinetics-Sounds have a unit in \texttt{pool4} which shows high
preference for the class. For the vision subnetwork the preference is
determined by ranking all images by their unit activation, and
retaining the top 5; if 4 out of these 5 images correspond to
one class, then that class is deemed to have a high-preference for the
unit (a similar procedure is carried out for the audio subnetwork
using spectrograms).  Our trained vision and audio networks have
high-preference units for 10 and 11 out of a possible 34 action
classes, respectively, compared to 1 and 1 for the random vision and
audio networks.  Furthermore, if the threshold for deeming a unit to
be high-preference is reduced to 3, our trained vision and audio
subnetworks cover 23 and 20 classes, respectively, compared to the 4
and 3 of a random network, respectively.  These results confirm that
our network has indeed learnt semantic features.

Furthermore,
Figure \ref{fig:tsne} shows the comparison between the trained and the
non-trained (\ie network with random weights) $L^3$-Net representations
for the visual and the audio modalities, on the Kinetics-Sounds dataset,
using the t-SNE visualization \cite{Van-der-Maaten08}.
It is clear that training for the audio-visual correspondence task
produces representations that have a semantic meaning, as videos containing
the same action classes
often cluster together, while the random network's representations do not
exhibit any clustering.
There is still a fair amount of confusion in the representations,
but this is expected as no class-level supervision is provided
and classes can be very alike.
For example, an organ and a piano are quite visually similar as they
contain keyboards, and the visual difference between a bass guitar and an
acoustic guitar is also quite fine; these similarities are reflected
in the closeness or overlap of respective clusters in
Figure \ref{fig:tsne}(c)
(\eg as noted earlier, ``fingerpicking'' is mostly associated with acoustic guitars).

\tabNMI

We also evaluate the quality of the $L^3$-Net embeddings by clustering them
with k-means into 64 clusters and reporting the Normalized Mutual Information
(NMI) score between the clusters and the ground truth action classes.
Results in Table \ref{tab:nmi} confirm the emergence of semantics as
the $L^3$-Net embeddings outperform the best random baselines by 50-100\%.

The t-SNE visualization also shows some interesting features,
such as the ``typing'' class being divided into two clusters in the visual
domain. Further investigation reveals that all frames in one cluster
show both a keyboard and hands, while the second cluster contains
much fewer hands. Separating these two cases can be a good indication of
whether the typing action is happening at the moment captured by the
(frame, 1 second sound clip) pair, and thus whether the typing sound
is expected to be heard. Furthermore, we found that the ``typing''
audio samples appear in three clusters -- the two fairly pure clusters
(outlined in Figure \ref{fig:tsne}(a)) correspond to strong typing sounds
and talking while typing, respectively, and the remaining cluster,
which is very impure and intermingled with other action classes,
mostly corresponds to silence and background noise.

\figUnitsAmap
\figTsne

\section{Discussion}

We have shown that the network trained for the AVC task achieves
superior results on sound classification to recent methods
that pre-train and fix the visual networks (one each for ImageNet and Scenes),
and we conjecture that the reason for this is that the additional freedom
of the visual network allows the learning to better take advantage of the 
opportunities offered by the variety of visual information in the video
(rather than be restricted to seeing only through the eyes of the pre-trained
network).
Also, the visual features that emerge from the $L^3$-Net
are on par with the state-of-the-art among self-supervised approaches.
Furthermore,
it has been demonstrated that the network automatically learns,
in both modalities,
fine-grained distinctions such as bass versus acoustic guitar or
saxophone versus clarinet.

The localization visualization results are reminiscent of the classic
highlighted pixels in~\cite{Kidron05}, except in our case we do not just learn
the few pixels that move (concurrent with the sound) but instead are able
to learn extended regions corresponding to the instrument.

We motivated this work by considering {\em correlation} of video and audio
events. However, we believe there is additional information in \emph{concurrency}
of the two streams, as concurrency is stronger than correlation
because the events need to be
synchronised (of course, if events are concurrent then they will
correlate, but not vice versa). Training for concurrency will require
video (multiple frames) as input, rather than a single video frame,
but it would be interesting to explore what more is gained from this
stronger condition.

In the future,
it would be interesting to learn
from the recently released large dataset of videos curated according to
{\em audio}, rather than visual, events~\cite{Gemmeke17} and see what subtle
visual semantic categories are discovered.
{\small
\bibliographystyle{ieee}
\bibliography{bib/shortstrings,bib/vgg_local,bib/vgg_other,bib/to_add}
}

\appendix
\section{Kinetics-Sounds}
\label{sec:kineticssounds}
The 34 action classes taken from the Kinetics dataset \cite{Kay17} to form
the Kinetics-Sounds dataset \ref{sec:datasets} are:
blowing nose,
bowling,
chopping wood,
ripping paper,
shuffling cards,
singing,
tapping pen,
typing,
blowing out,
dribbling ball,
laughing,
mowing the lawn by pushing lawnmower,
shoveling snow,
stomping,
tap dancing,
tapping guitar,
tickling,
fingerpicking,
patting,
playing accordion,
playing bagpipes,
playing bass guitar,
playing clarinet,
playing drums,
playing guitar,
playing harmonica,
playing keyboard,
playing organ,
playing piano,
playing saxophone,
playing trombone,
playing trumpet,
playing violin,
playing xylophone.

\end{document}

%% file: floats.tex
\newcommand{\figTask}{
\begin{figure}[t]
\begin{center}
  \includegraphics[width=\linewidth]{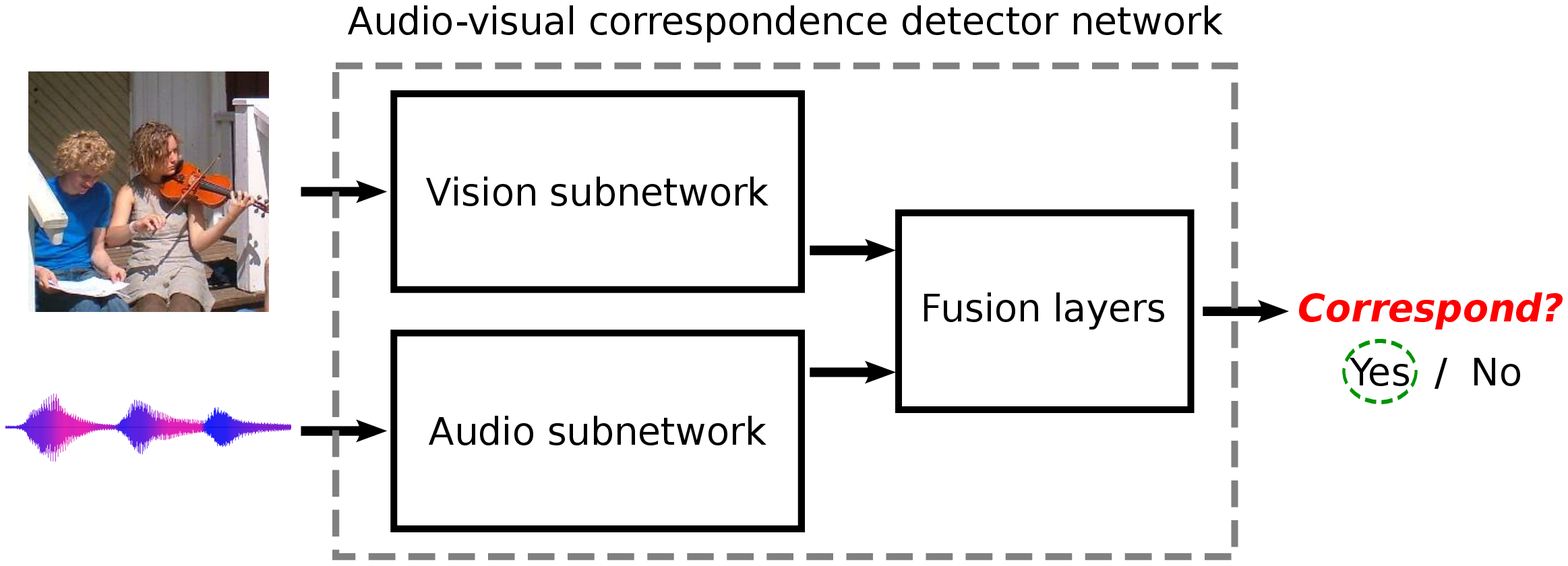}
\end{center}
   \caption{{\bf Audio-visual correspondence task (AVC)}.
A network should learn to determine whether a pair of
(video frame, short audio clip) correspond to each other or not.
Positives are (frame, audio) extracted from the same time of one video,
while negatives are a frame and audio extracted from different videos.
}
\label{fig:task}
\end{figure}
}

\newcommand{\figNet}{
\begin{figure}[t]
\vspace{-0.3cm}
\begin{center}
  \includegraphics[height=0.5\textheight]{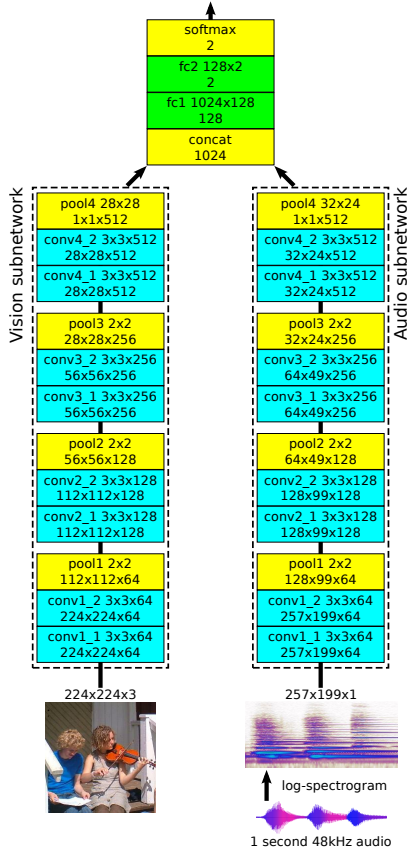}
\end{center}
\vspace{-0.2cm}
   \caption{{\bf $L^3$-Net architecture}.
Each blocks represents a single layer with text providing more information --
first row: layer name and parameters, second row: output feature map size.
Layers with a name prefix
\texttt{conv}, \texttt{pool}, \texttt{fc}, \texttt{concat}, \texttt{softmax}
are convolutional, max-pooling, fully connected, concatenation and softmax
layers, respectively.
The listed parameters are:
conv -- kernel size and number of channels,
pooling -- kernel size,
fc -- size of the weight matrix.
The stride of pool layers is equal to the kernel size and there is no padding.
Each convolutional layer is followed by batch normalization \cite{Ioffe15}
and a ReLU nonlinearity, and the first fully connected layer (\texttt{fc1})
is followed by ReLU.
}
\label{fig:net}
\vspace{-0.3cm}
\end{figure}
}

\newcommand{\tabAvc}{
\begin{table}[t]
\begin{center}
\small
\begin{tabular}{lcc}
    Method & Flickr-SoundNet & Kinetics-Sounds \\
    \hline\hline
    Supervised direct & -- & 65\% \\ %
    Supervised pretraining & -- & 74\% \\
    $L^3$-Net & 78\% & 74\%
\end{tabular}
\end{center}
\vspace{-0.2cm}
    \caption{{\bf Audio-visual correspondence (AVC) results.}
Test set accuracy on the AVC task for the $L^3$-Net,
and the %
two supervised baselines on the labelled
Kinetics-Sounds dataset.
The number of positives and negatives is the same, so chance gets 50\%.
All methods are trained on the training set of the respective datasets.
}
\label{tab:avc}
\end{table}
}

\newcommand{\tabSound}{
\begin{table}[t]
\begin{center}
\small
\hspace*{-0.3cm}
\begin{tabular}{c@{~~~}c}
(a) ESC-50 & (b) DCASE \\
\begin{adjustbox}{valign=t}
\begin{tabular}{lc}
    Method & Accuracy \\
    \hline\hline
    SVM-MFCC \cite{Piczak15} & 39.6\% \\
    Autoencoder \cite{Aytar16} & 39.9\% \\
    Random Forest \cite{Piczak15} & 44.3\% \\
    Piczak ConvNet \cite{Piczak15a} & 64.5\% \\
    SoundNet \cite{Aytar16} & 74.2\% \\
    \hline
    Ours random & 62.5\% \\
    Ours & {\bf 79.3\%} \\
    \hline
    \emph{Human perf.\ \cite{Piczak15}} & \emph{81.3\%}
\end{tabular}
\end{adjustbox}
&
\begin{adjustbox}{valign=t}
\begin{tabular}{lc}
    Method & Accuracy \\
    \hline\hline
    RG \cite{Rakotomamonjy15} & 69\% \\
    LTT \cite{Li13a} & 72\% \\
    RNH \cite{Roma13} & 77\% \\
    Ensemble \cite{Stowell15} & 78\% \\
    SoundNet \cite{Aytar16} & 88\% \\
    \hline
    Ours random & 85\% \\
    Ours & {\bf 93\%}
\end{tabular}
\end{adjustbox}
\end{tabular}
\end{center}
\vspace{-0.2cm}
    \caption{{\bf Sound classification.}
``Ours random'' is an additional baseline which shows the performance
of our network without $L^3$-training.
Our $L^3$-training sets the new state-of-the-art by a large margin
on both benchmarks.
}
\label{tab:sound}
\end{table}
}

\newcommand{\tabImageNet}{
\begin{table}[t]
\begin{center}
\small
\begin{tabular}{lc}%
    Method & Top 1 accuracy \\ %
    \hline\hline
    Random & 18.3\% \\ %
    Pathak \etal \cite{Pathak16} & 22.3\% \\ %
    Kr\"ahenb\"uhl \etal \cite{Krahenbuhl15} & 24.5\% \\ %
    Donahue \etal \cite{Donahue17} & 31.0\% \\ %
    Doersch \etal \cite{Doersch15} & 31.7\% \\ %
    Zhang \etal \cite{Zhang16} (init: \cite{Krahenbuhl15}) & 32.6\% \\ %
    Noroozi and Favaro \cite{Noroozi16} & 34.7\% \\
    \hline
    Ours random & 12.9\% \\ %
    Ours & 32.3\% \\ %
\end{tabular}
\end{center}
\vspace{-0.2cm}
    \caption{{\bf Visual classification on ImageNet.}
Following \cite{Zhang16}, our features are evaluated by training a linear
classifier on the ImageNet training set and measuring the classification
accuracy on the validation set.
For more details and discussions see Section \ref{sec:imagenet}.
All performance numbers apart from ours are provided by authors of \cite{Zhang16},
showing only the best performance for each method over all parameter choices
(\eg Donahue \etal \cite{Donahue17} achieve 27.1\% instead of 31.0\% when taking
features from \texttt{pool5} instead of \texttt{conv3}).
}
\label{tab:imagenet}
\vspace{-0.3cm}
\end{table}
}

\newcommand{\figUnitsV}{
\def\unitsW{0.1\linewidth}
\def\fntsz{\scriptsize}
\begin{figure*}[p]
\begin{center}
\setlength{\tabcolsep}{2pt}
\begin{tabular}{ccccccccc}
  \fntsz Fingerpicking &
  \fntsz Lawn mowing &
  \fntsz P.\ accordion &
  \fntsz P.\ bass guitar &
  \fntsz P.\ saxophone &
  \fntsz Typing &
  \fntsz Bowling &
  \fntsz P.\ clarinet &
  \fntsz P.\ organ
  \\
  \includegraphics[width=\unitsW]{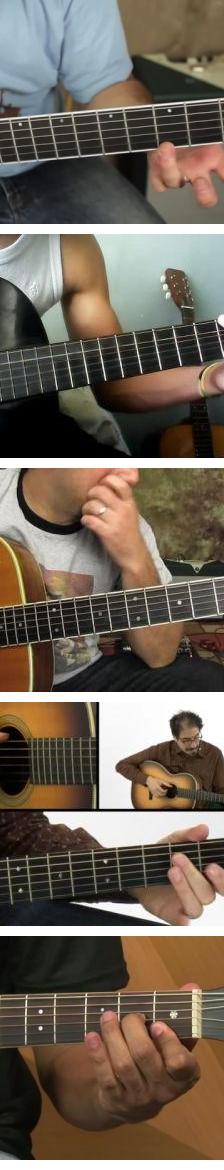} &
  \includegraphics[width=\unitsW]{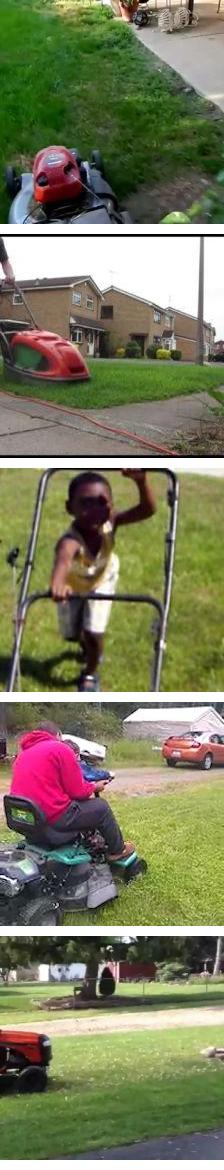} &
  \includegraphics[width=\unitsW]{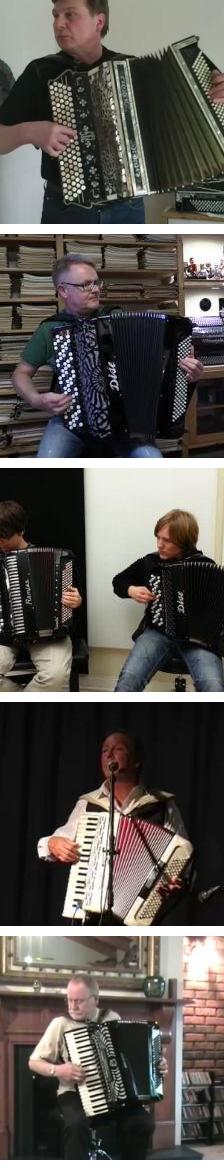} &
  \includegraphics[width=\unitsW]{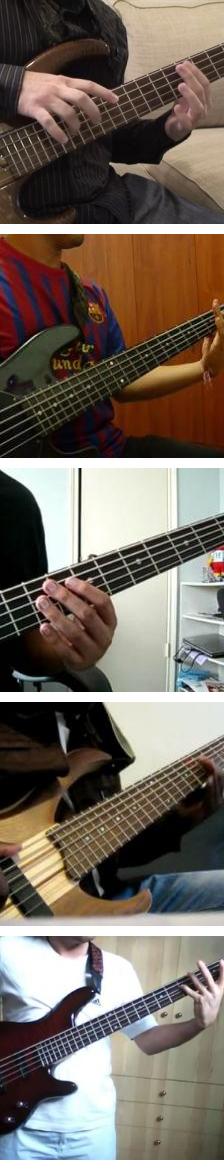} &
  \includegraphics[width=\unitsW]{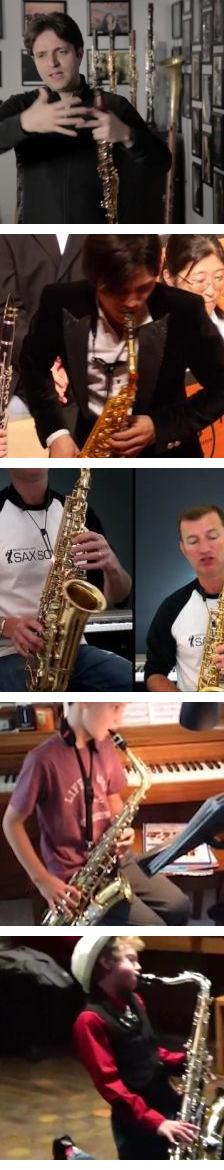} &
  \includegraphics[width=\unitsW]{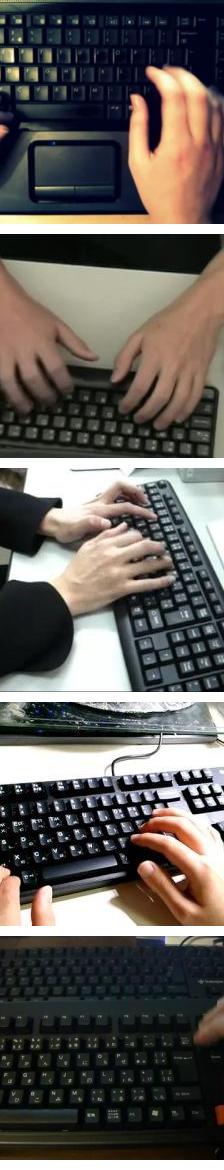} &
  \includegraphics[width=\unitsW]{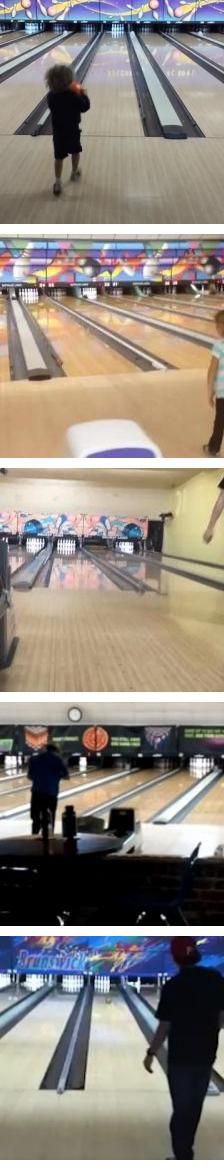} &
  \includegraphics[width=\unitsW]{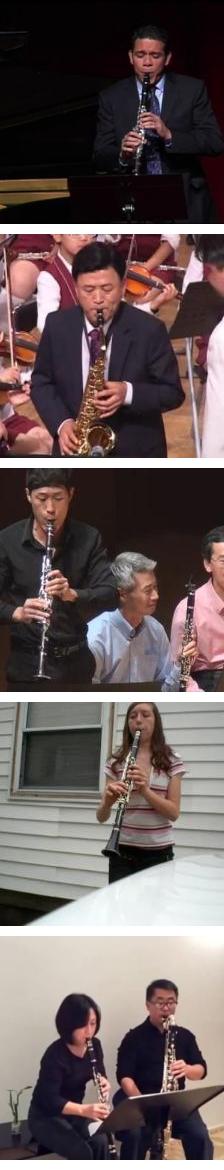} &
  \includegraphics[width=\unitsW]{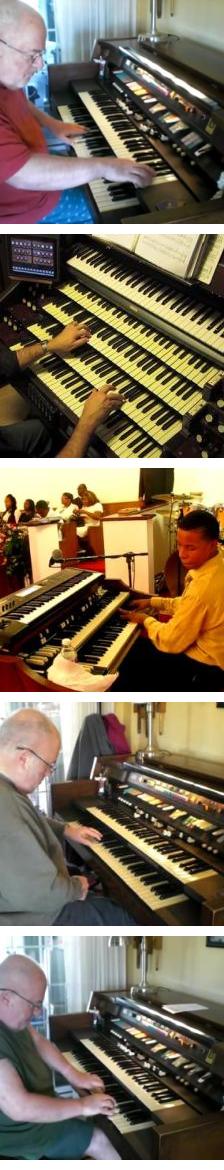}
\end{tabular}
\end{center}
\vspace{-0.4cm}
   \caption{{\bf Learnt visual concepts (Kinetics-Sounds).}
Each column shows five images that most activate
a particular unit of the 512 in \texttt{pool4} for the vision subnetwork.
Note that these features do not take sound as input.
Videos come from the Kinetics-Sounds test set and the network was trained
on the Kinetics-Sounds train set.
The top row shows the dominant action label for the unit (``P.'' stands for ``playing'').
}
\label{fig:unitsV}
\end{figure*}
}

\newcommand{\figUnitsVmap}{
\def\unitsW{0.1\linewidth}
\def\fntsz{\scriptsize}
\begin{figure*}[p]
\begin{center}
\setlength{\tabcolsep}{2pt}
\begin{tabular}{ccccccccc}
  \fntsz Fingerpicking &
  \fntsz Lawn mowing &
  \fntsz P.\ accordion &
  \fntsz P.\ bass guitar &
  \fntsz P.\ saxophone &
  \fntsz Typing &
  \fntsz Bowling &
  \fntsz P.\ clarinet &
  \fntsz P.\ organ
  \\
  \includegraphics[width=\unitsW]{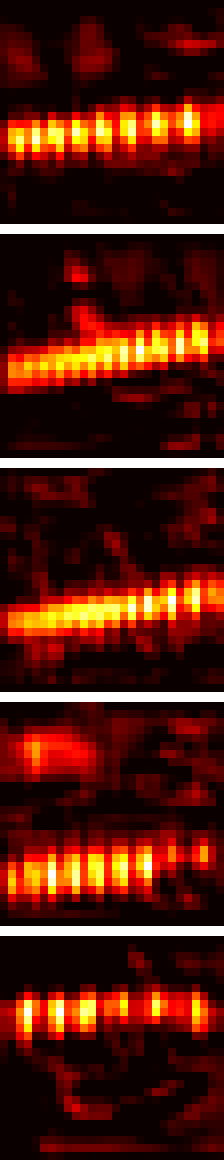} &
  \includegraphics[width=\unitsW]{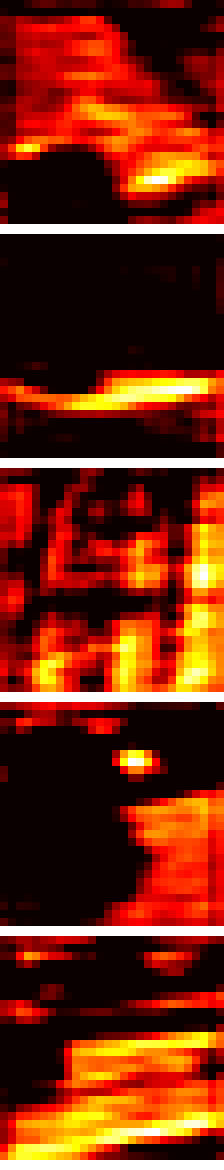} &
  \includegraphics[width=\unitsW]{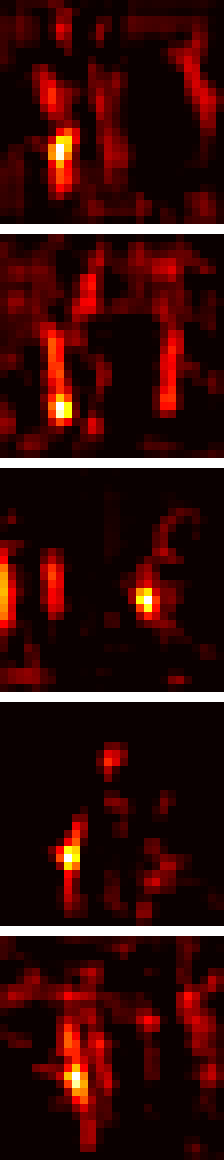} &
  \includegraphics[width=\unitsW]{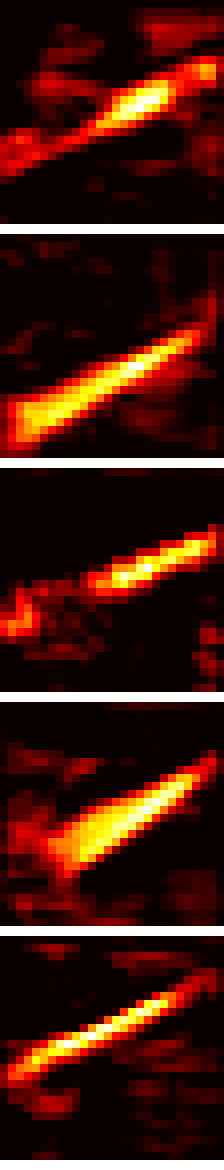} &
  \includegraphics[width=\unitsW]{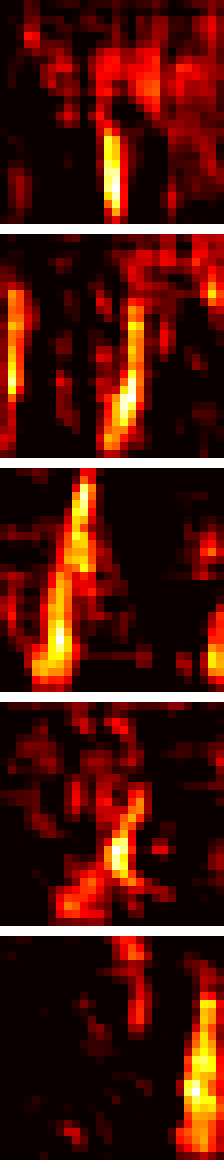} &
  \includegraphics[width=\unitsW]{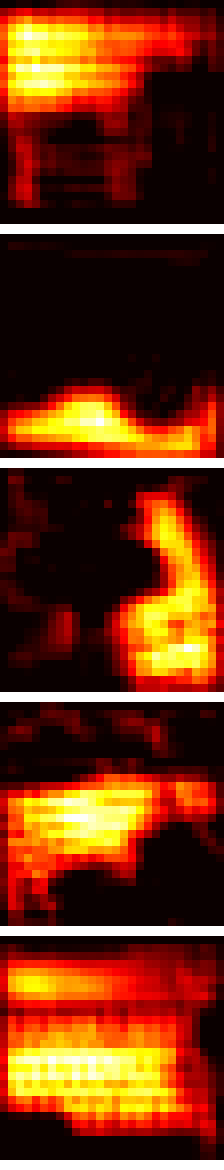} &
  \includegraphics[width=\unitsW]{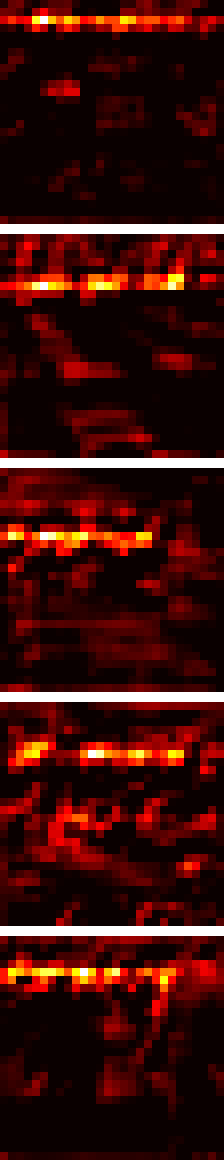} &
  \includegraphics[width=\unitsW]{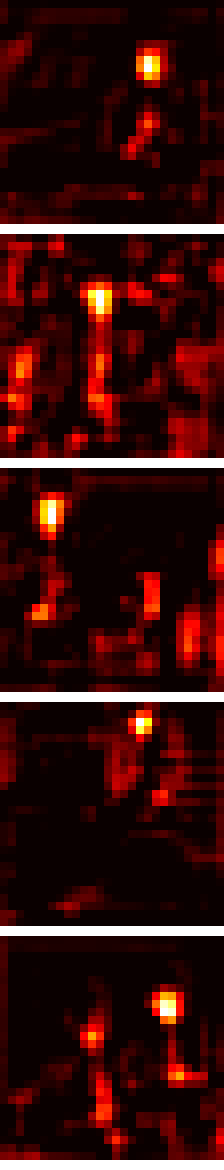} &
  \includegraphics[width=\unitsW]{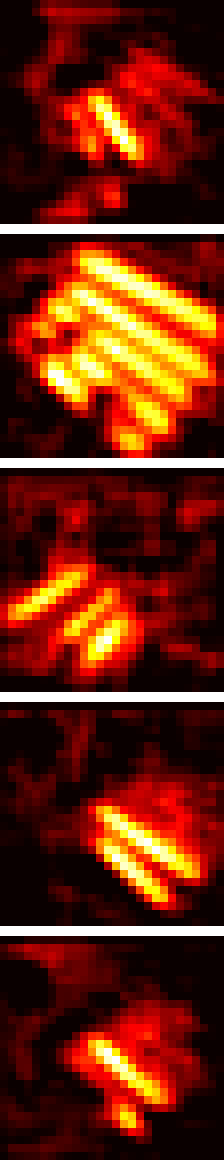}
\end{tabular}
\end{center}
\vspace{-0.4cm}
   \caption{{\bf Visual semantic heatmap (Kinetics-Sounds).}
Examples correspond to the ones in Figure \ref{fig:unitsV}.
A semantic heatmap is obtained as a slice of activations from
\texttt{conv4\_2} of the vision subnetwork that corresponds to the same unit from
\texttt{pool4} as in Figure \ref{fig:unitsV},
\ie the unit that responds highly to the class in question.
}
\label{fig:unitsVmap}
\end{figure*}
}

\newcommand{\figUnitsA}{
\def\unitsW{0.1\linewidth}
\def\fntsz{\scriptsize}
\begin{figure*}[t]
\vspace{-0.3cm}
\begin{center}
\setlength{\tabcolsep}{2pt}
\begin{tabular}{ccccccccc}
  \fntsz Fingerpicking &
  \fntsz Lawn mowing &
  \fntsz P.\ accordion &
  \fntsz P.\ bass guitar &
  \fntsz P.\ saxophone &
  \fntsz Typing &
  \fntsz P.\ xylophone &
  \fntsz Tap dancing &
  \fntsz Tickling
  \\
  \includegraphics[width=\unitsW]{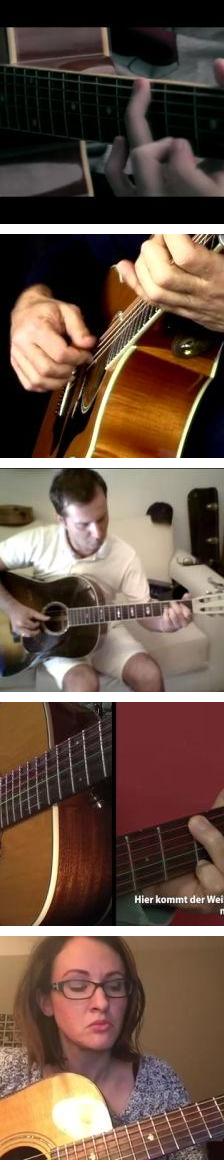} &
  \includegraphics[width=\unitsW]{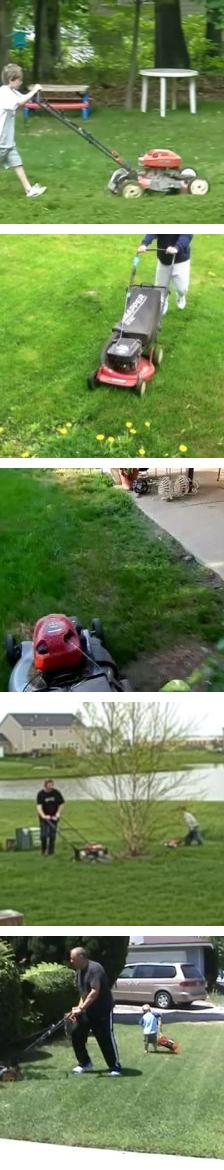} &
  \includegraphics[width=\unitsW]{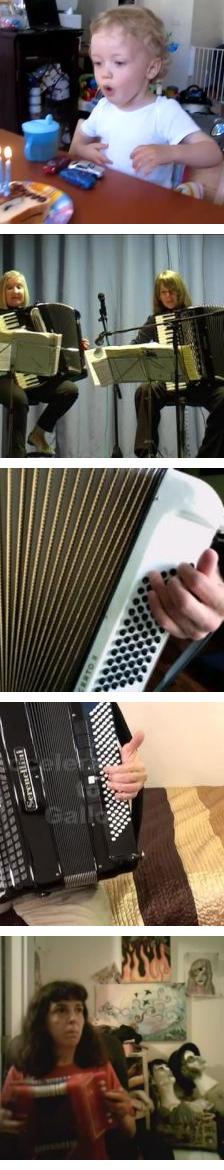} &
  \includegraphics[width=\unitsW]{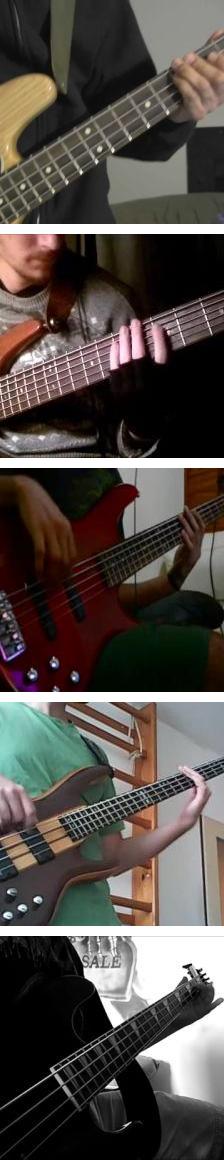} &
  \includegraphics[width=\unitsW]{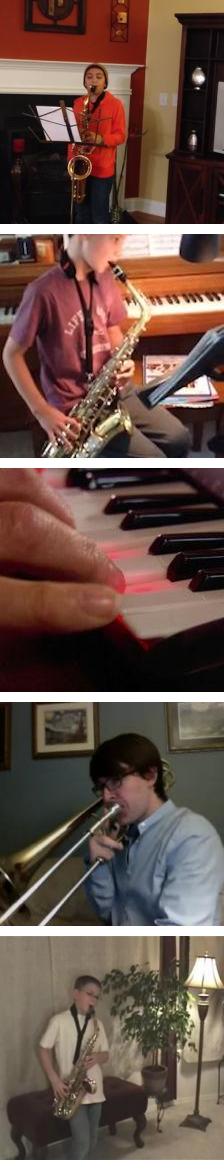} &
  \includegraphics[width=\unitsW]{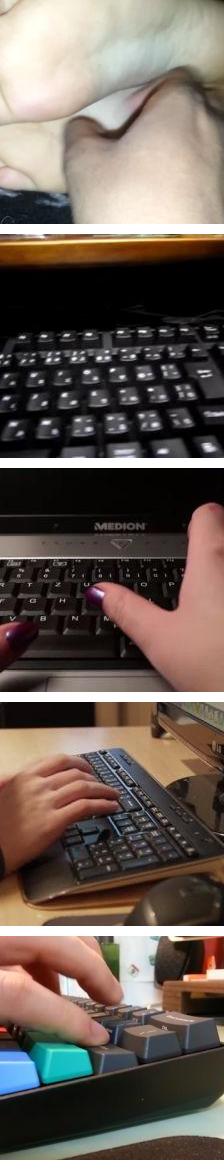} &
  \includegraphics[width=\unitsW]{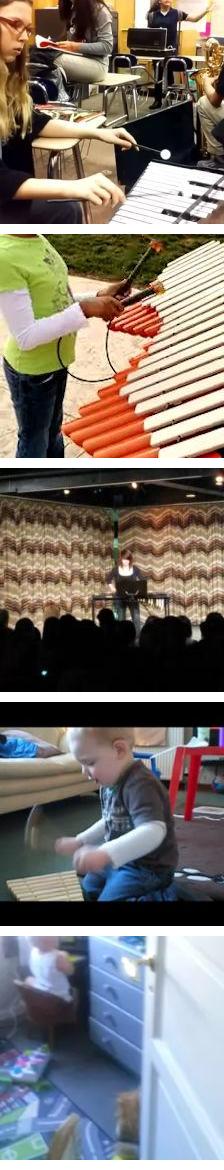} &
  \includegraphics[width=\unitsW]{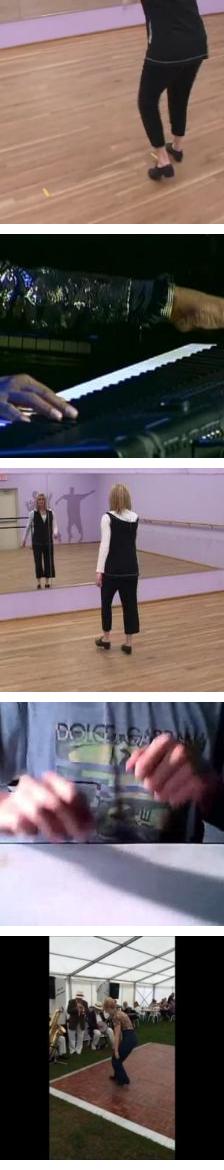} &
  \includegraphics[width=\unitsW]{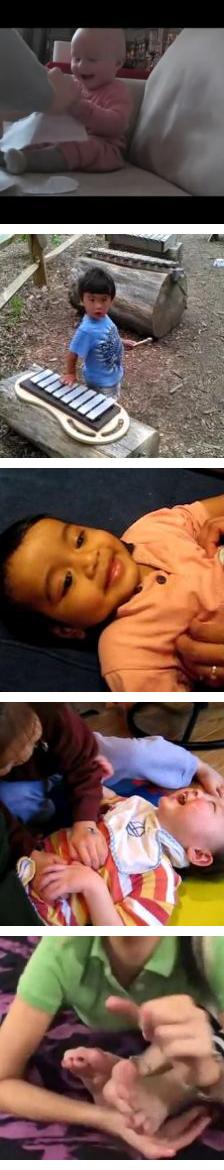}
\end{tabular}
\end{center}
\vspace{-0.4cm}
   \caption{{\bf Learnt audio concepts (Kinetics-Sounds).}
Each column shows five sounds that most activate
a particular unit in \texttt{pool4} of the audio subnetwork.
Purely for visualization purposes, as it is hard to display sound,
the frame of the video that is aligned with the sound is shown instead
of the actual sound form,
but we stress that no vision is used in this experiment.
Videos come from the Kinetics-Sounds test set and the network was trained
on the Kinetics-Sounds train set.
The top row shows the dominant action label for the unit (``P.'' stands for ``playing'').
}
\label{fig:unitsA}
\vspace{-0.3cm}
\end{figure*}
}

\newcommand{\figUnitsAmap}{
\def\unitsW{0.11\linewidth}
\def\fntsz{\scriptsize}
\begin{figure}[t]
\begin{center}
\setlength{\tabcolsep}{2pt}
\begin{tabular}{c@{~}c@{~~~}c@{~}c@{~~~}c@{~}c@{~~~}c@{~}c@{~~~}c@{~}c}
  \multicolumn{2}{c}{\fntsz Fingerpicking ~~~} &
  \multicolumn{2}{c}{\fntsz Lawn mowing ~~~} &
  \multicolumn{2}{c}{\fntsz P.\ bass guitar ~~~} &
  \multicolumn{2}{c}{\fntsz Tap dancing ~~~}
  \\
  \includegraphics[width=\unitsW]{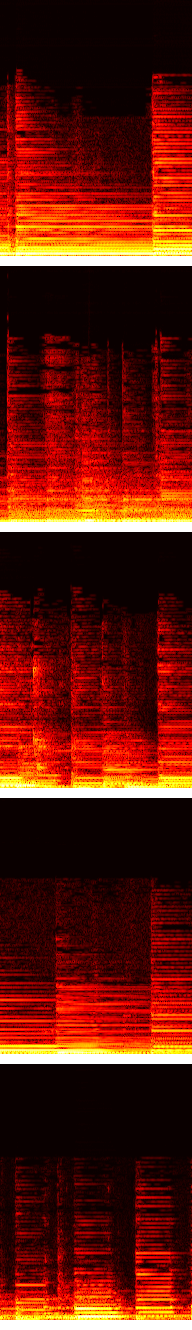} &
  \includegraphics[width=\unitsW]{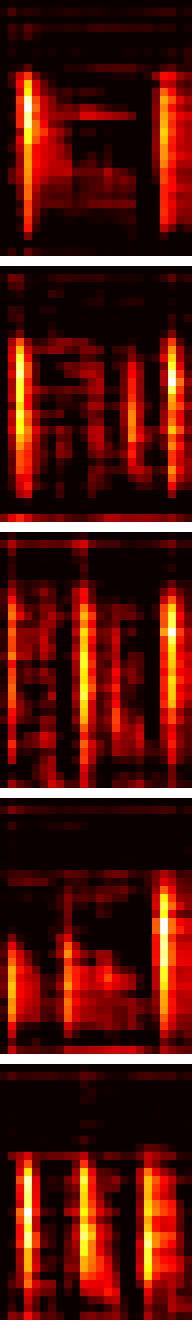} &
  \includegraphics[width=\unitsW]{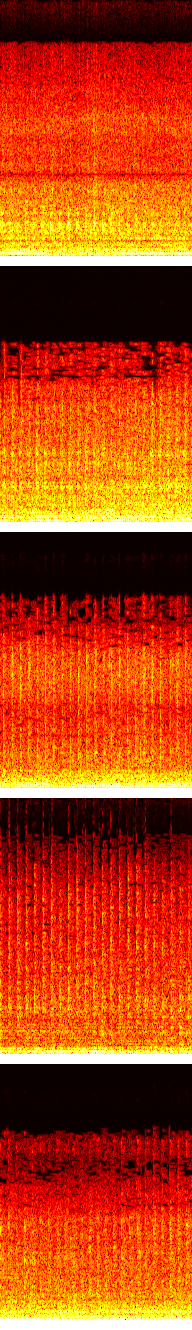} &
  \includegraphics[width=\unitsW]{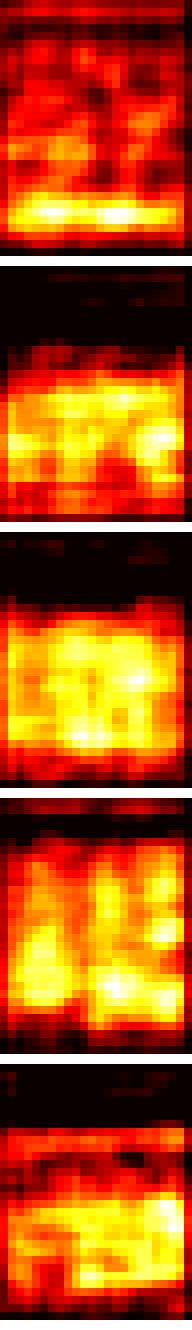} &
  \includegraphics[width=\unitsW]{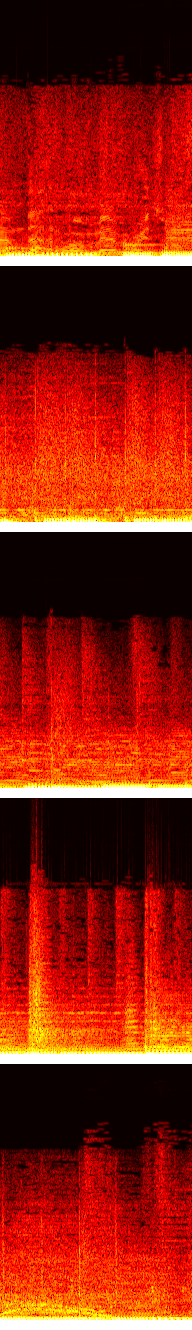} &
  \includegraphics[width=\unitsW]{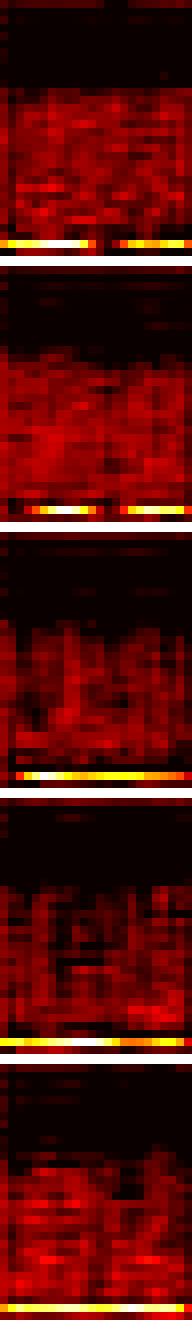} &
  \includegraphics[width=\unitsW]{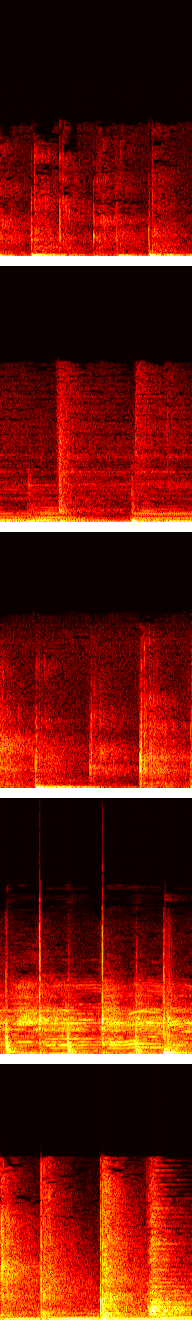} &
  \includegraphics[width=\unitsW]{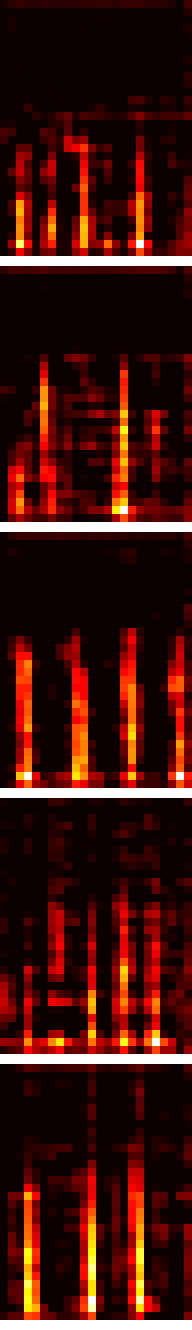}
\end{tabular}
\end{center}
   \caption{{\bf Audio semantic heatmaps (Kinetics-Sounds).}
Each pair of columns shows a single action class
(top, ``P.'' stands for ``playing''),
five log-spectrograms (left) and spectrogram semantic heatmaps (right)
for the class.
Horizontal and vertical axes correspond to the time and frequency dimensions,
respectively.
A semantic heatmap is obtained as a slice of activations of the unit from
\texttt{conv4\_2} of the audio subnetwork which shows preference for
the considered class.
}
\label{fig:unitsAmap}
\end{figure}
}

%% file: floats_supp.tex
\newcommand{\figUnitsVSoundNetA}{
\def\unitsW{0.144\linewidth}
\def\fntsz{\scriptsize}
\begin{figure*}[p]
\begin{center}
\setlength{\tabcolsep}{2pt}
\begin{tabular}{cccccc}
  \multicolumn{2}{c}{Outdoor} &
  \multicolumn{2}{c}{Concert} &
  \multicolumn{2}{c}{Outdoor sport}
  \\
  \includegraphics[width=\unitsW]{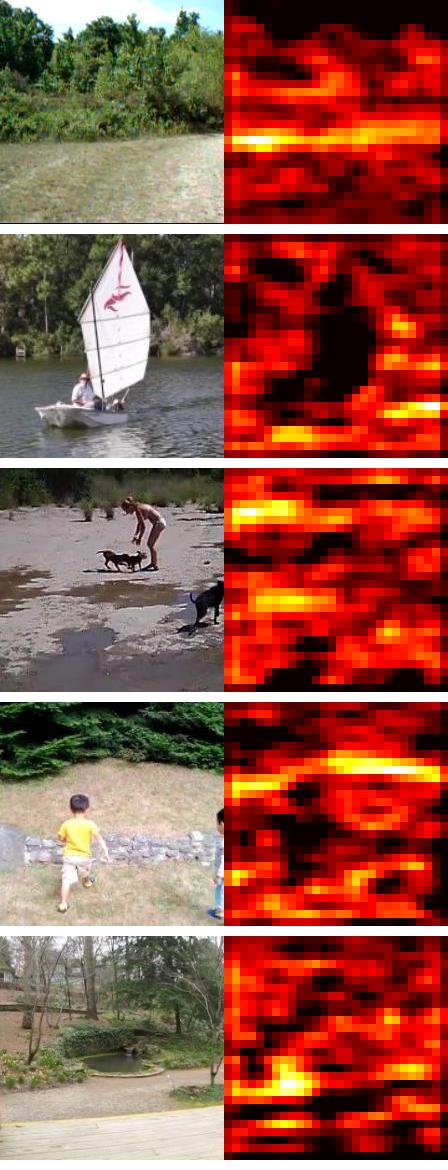} &
  \includegraphics[width=\unitsW]{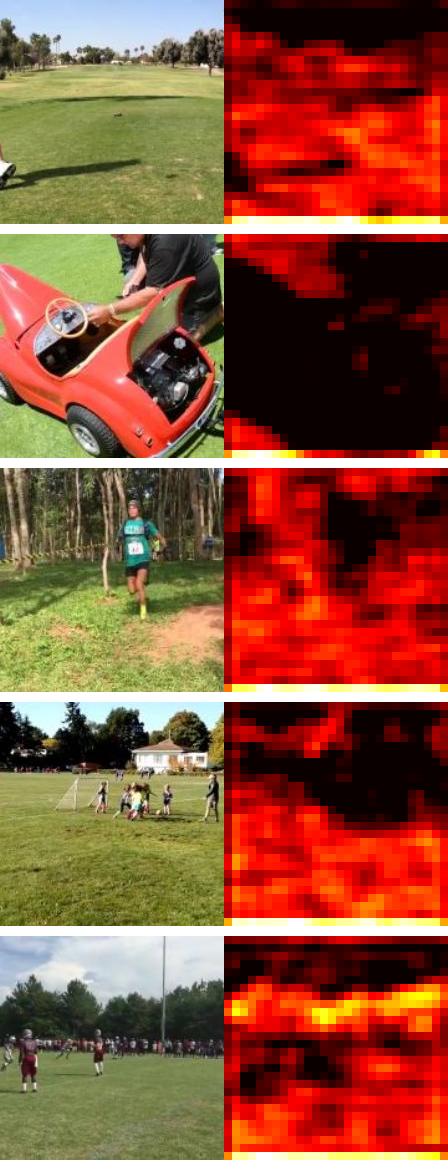} &
  \includegraphics[width=\unitsW]{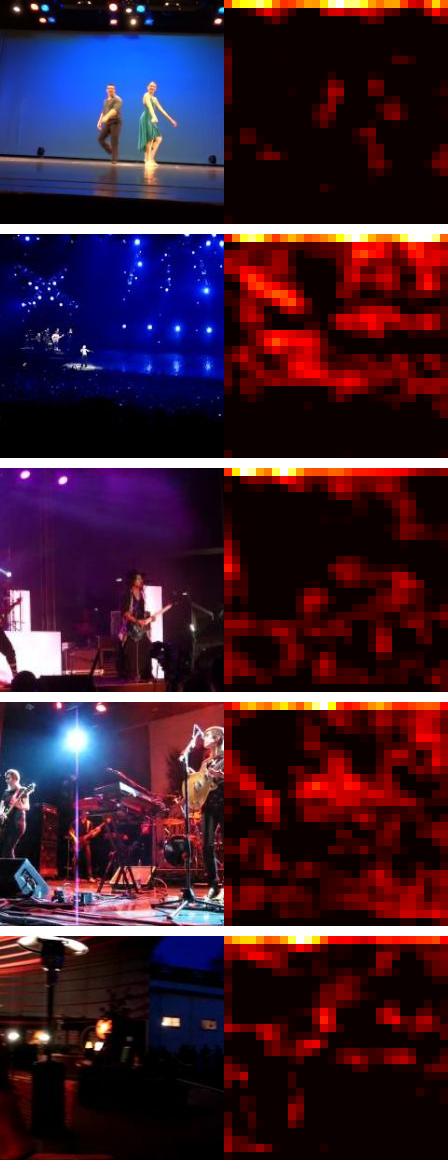} &
  \includegraphics[width=\unitsW]{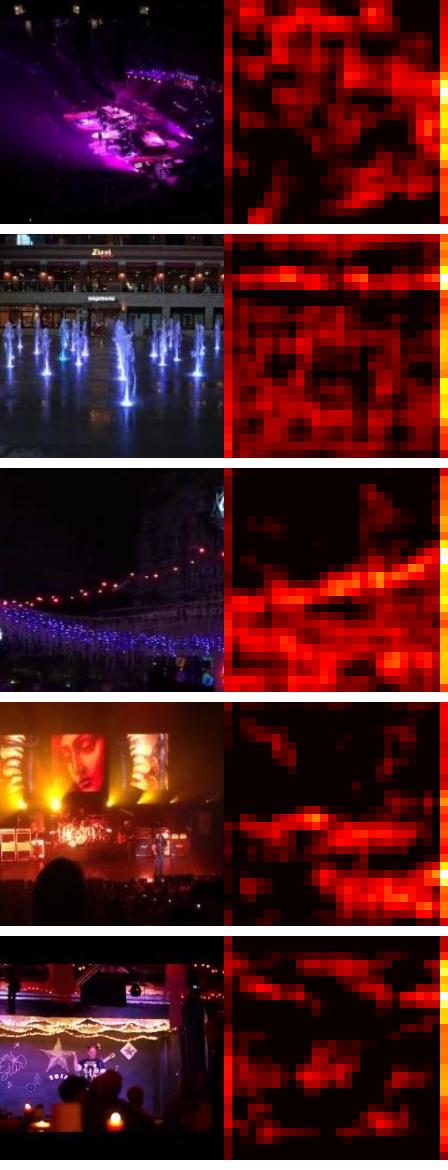} &
  \includegraphics[width=\unitsW]{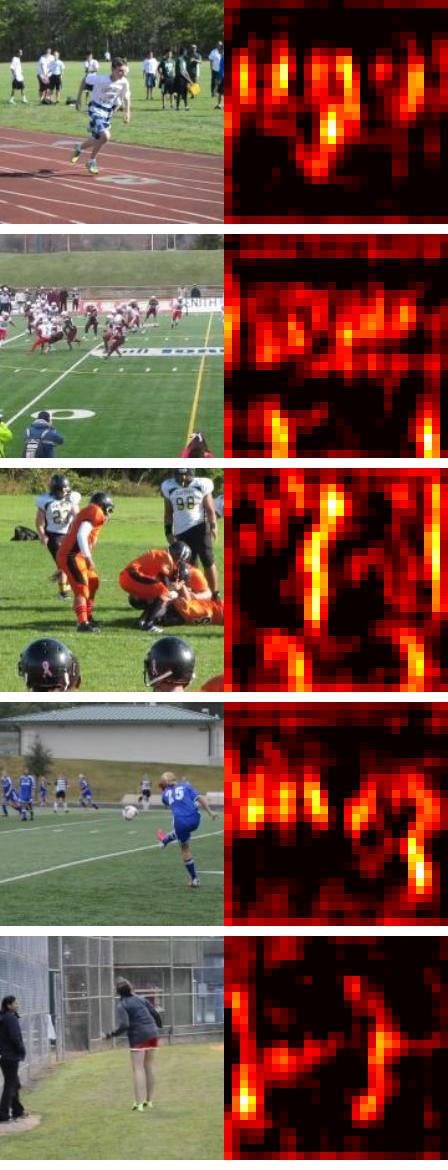} &
  \includegraphics[width=\unitsW]{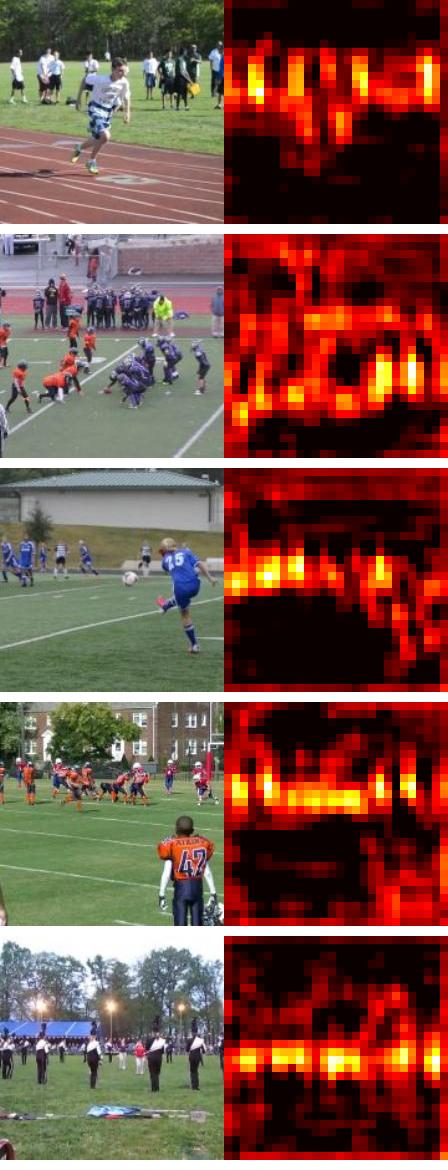}
  \\ \hline \\
  Cloudy sky &
  \multicolumn{2}{c}{Sky} &
  \multicolumn{2}{c}{Water surface} &
  Underwater
  \\
  \includegraphics[width=\unitsW]{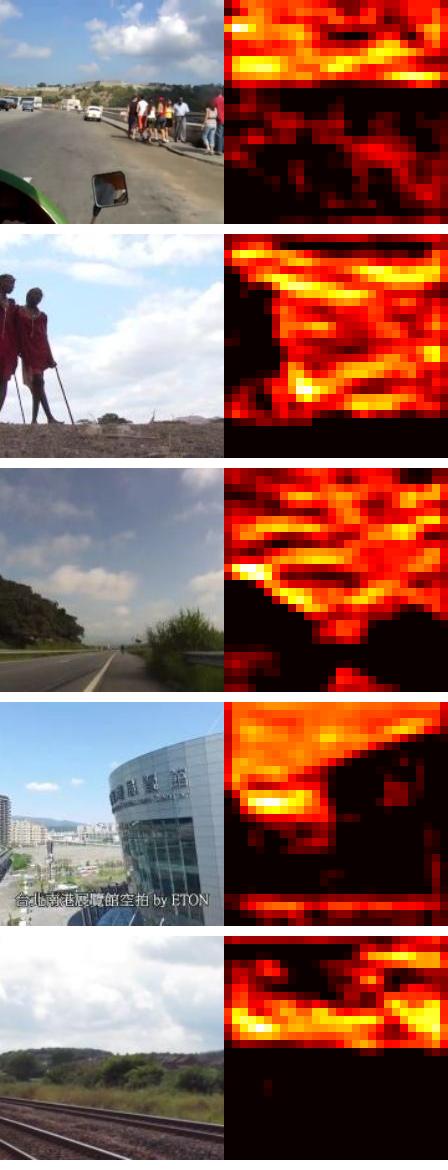} &
  \includegraphics[width=\unitsW]{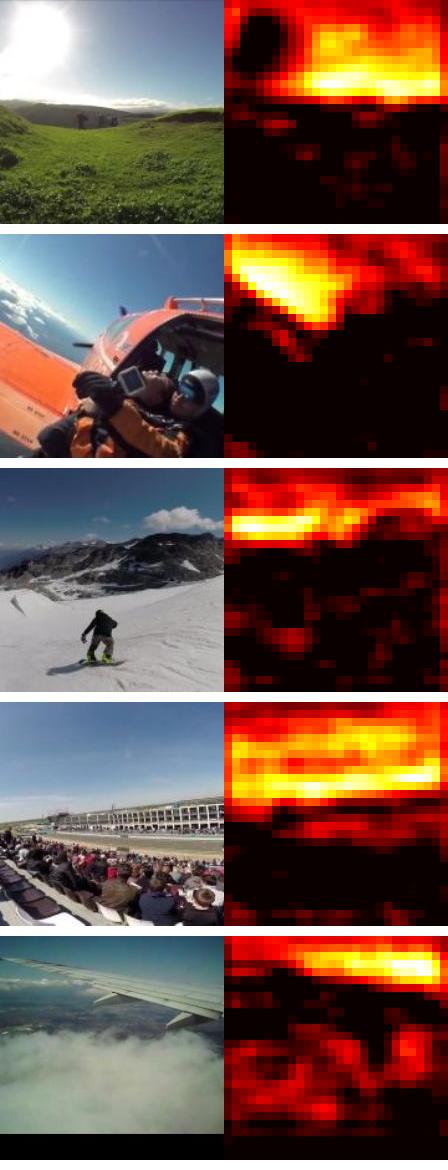} &
  \includegraphics[width=\unitsW]{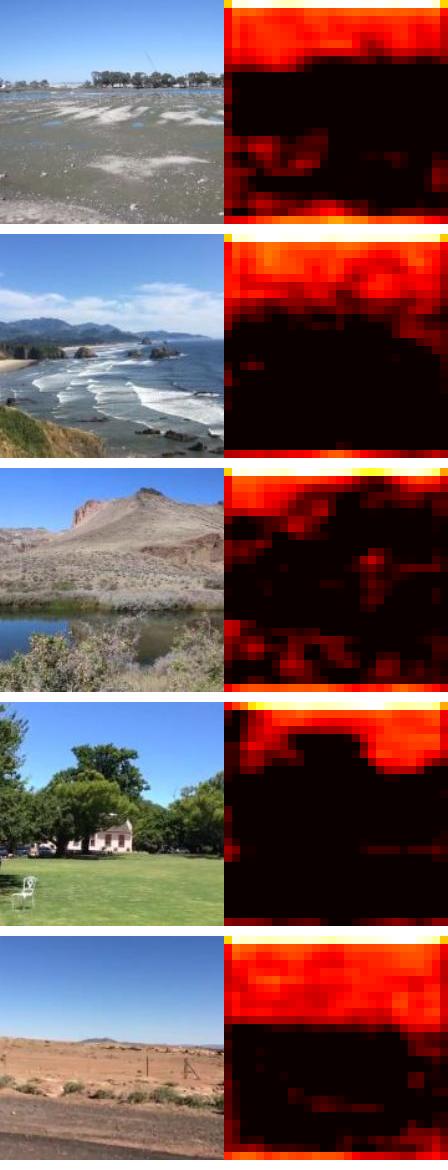} &
  \includegraphics[width=\unitsW]{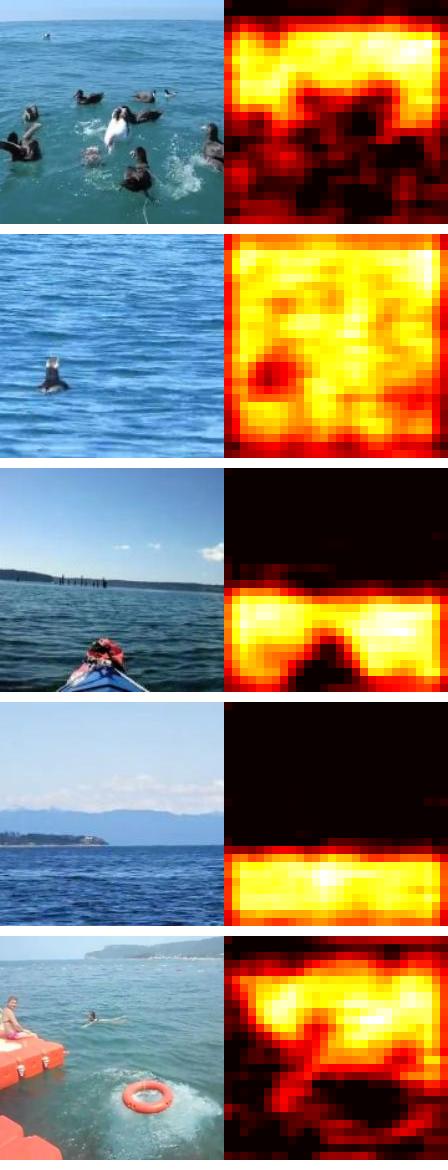} &
  \includegraphics[width=\unitsW]{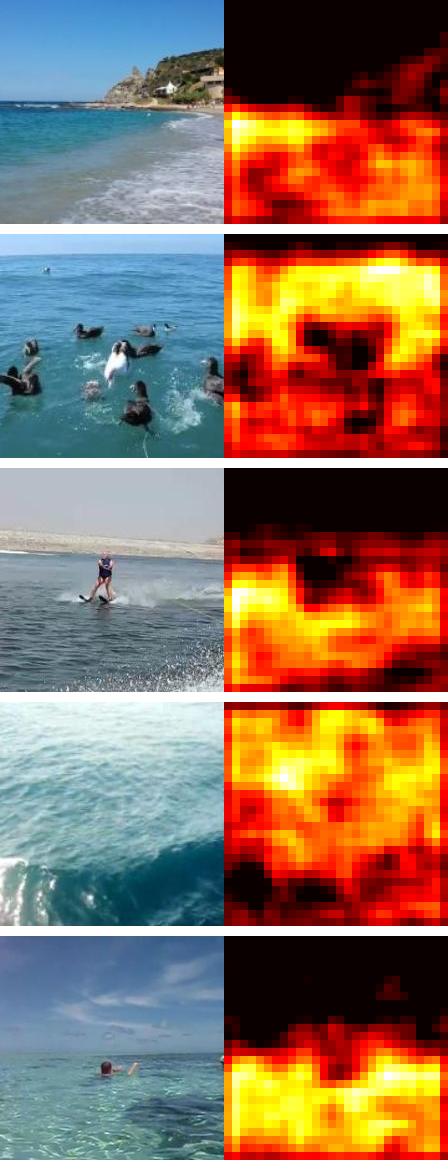} &
  \includegraphics[width=\unitsW]{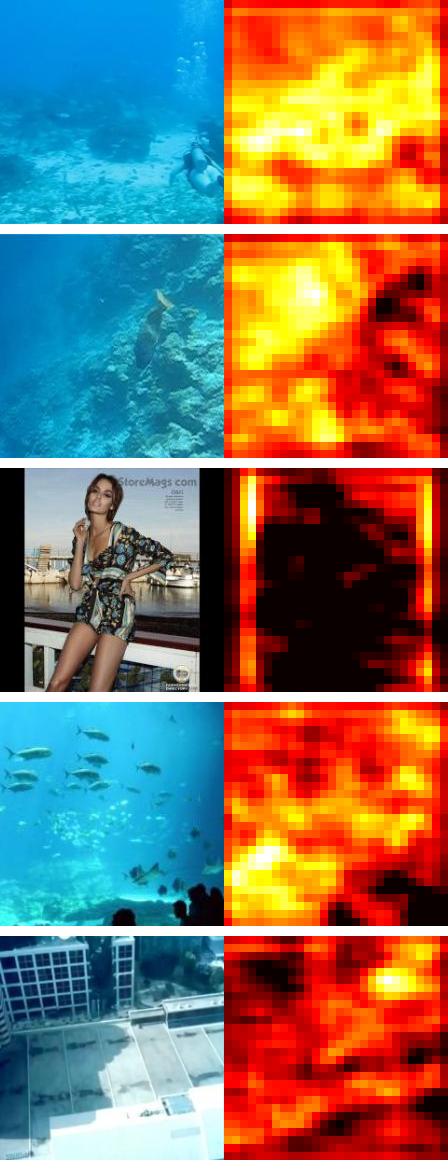}
  \\ \hline \\
  Horizon &
  Railway &
  \multicolumn{2}{c}{Crowd} &
  \multicolumn{2}{c}{Text}
  \\
  \includegraphics[width=\unitsW]{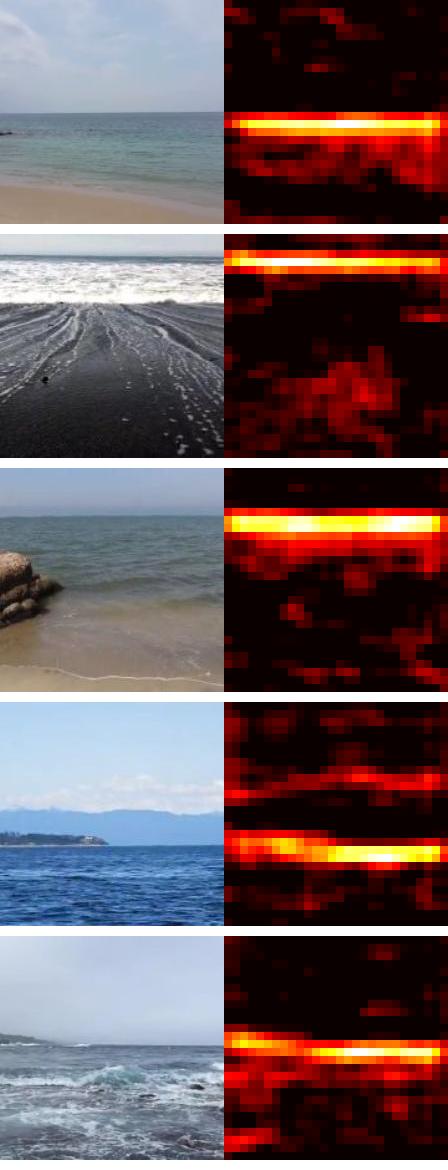} &
  \includegraphics[width=\unitsW]{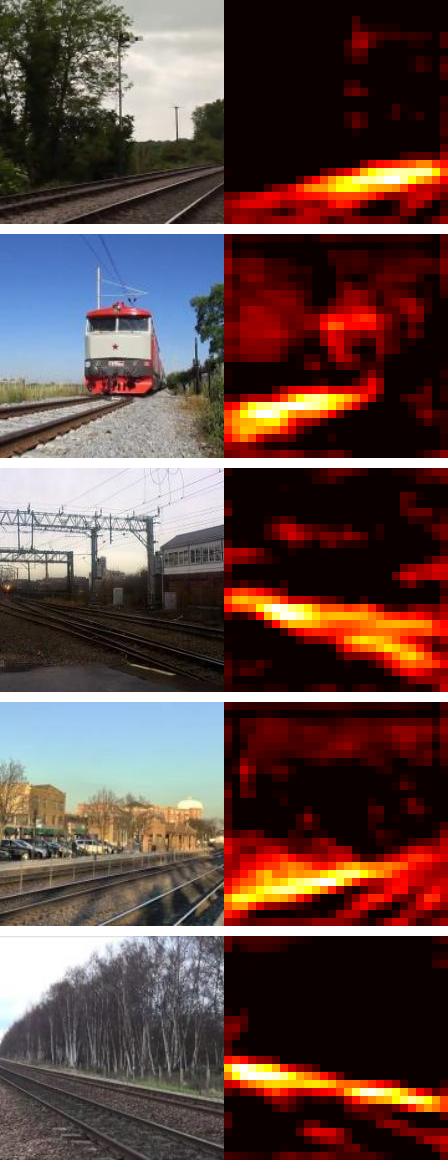} &
  \includegraphics[width=\unitsW]{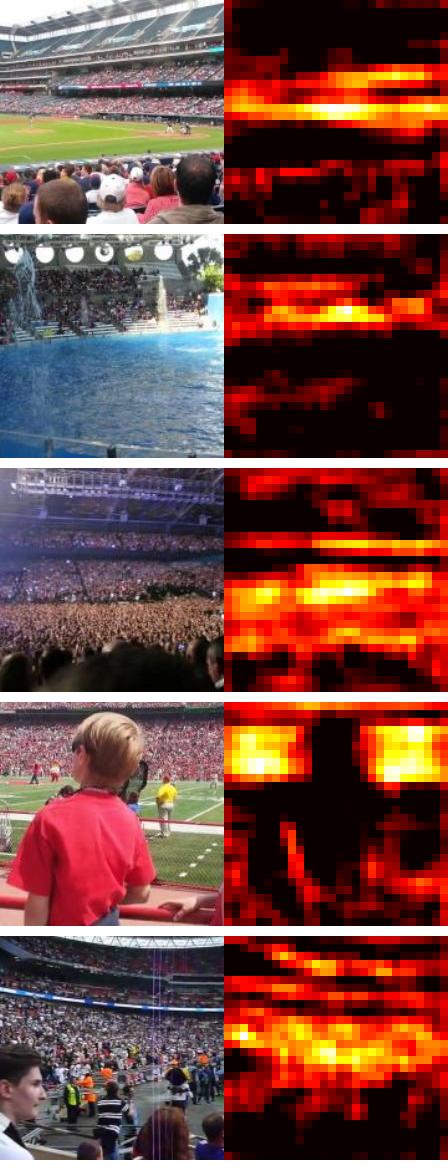} &
  \includegraphics[width=\unitsW]{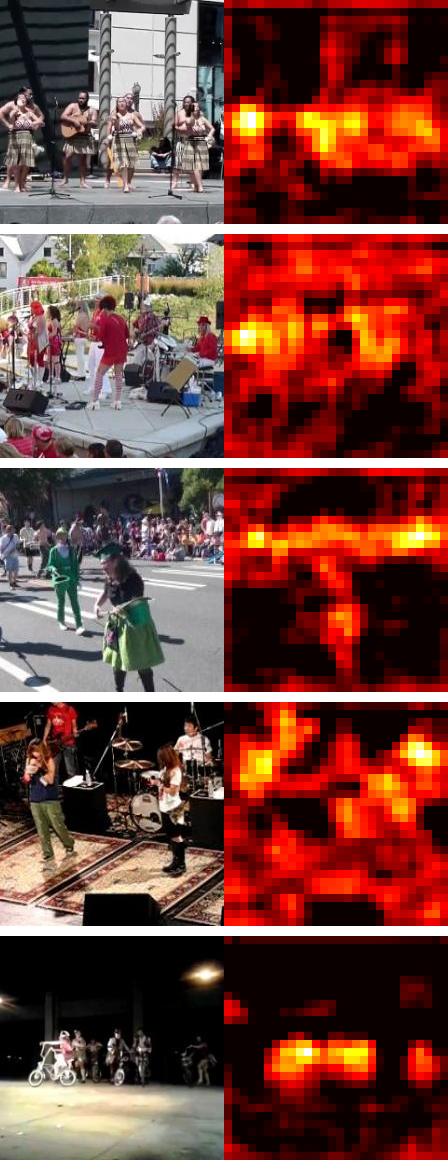} &
  \includegraphics[width=\unitsW]{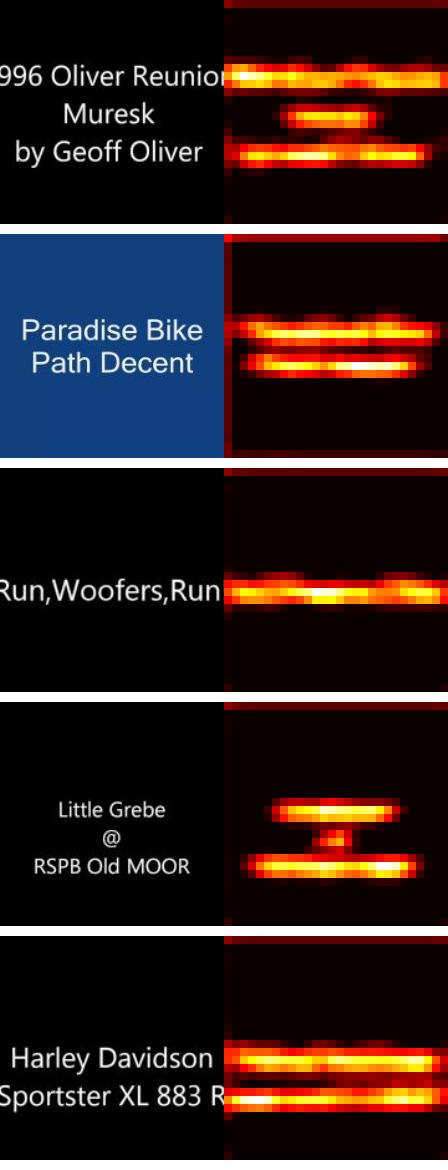} &
  \includegraphics[width=\unitsW]{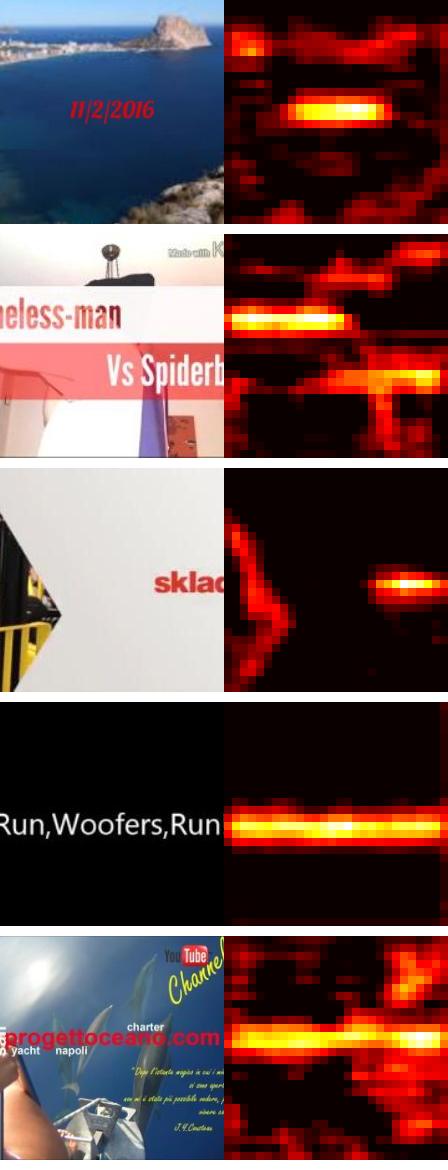}
\end{tabular}
\end{center}
\vspace{-0.4cm}
   \caption{{\bf Learnt visual concepts and semantic heatmaps (Flickr-SoundNet).}
Each mini-column shows five images that most activate a particular unit of the 512 in
\texttt{pool4} of the vision subnetwork, and the corresponding heatmap
(for more details see Figures \ref{fig:unitsV} and \ref{fig:unitsVmap}).
Column titles are a subjective names of concepts the units respond to.
}
\label{fig:unitsVSoundNetA}
\end{figure*}
}

\newcommand{\figUnitsVSoundNetB}{
\def\unitsW{0.19\linewidth}
\def\fntsz{\scriptsize}
\begin{figure*}[b]
\begin{center}
\setlength{\tabcolsep}{2pt}
\begin{tabular}{ccccc}
  Baby &
  Face &
  \multicolumn{2}{c}{Head} &
  Crowd
  \\
  \includegraphics[width=\unitsW]{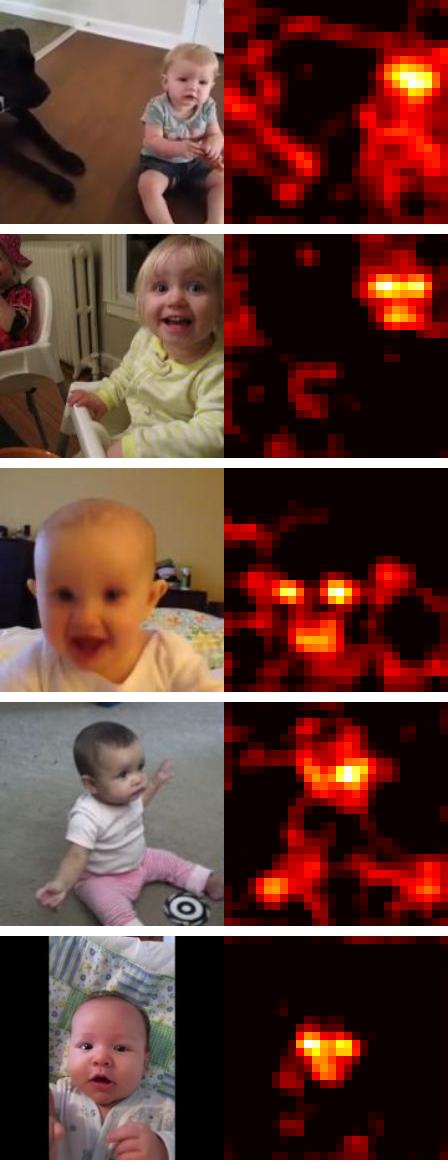} &
  \includegraphics[width=\unitsW]{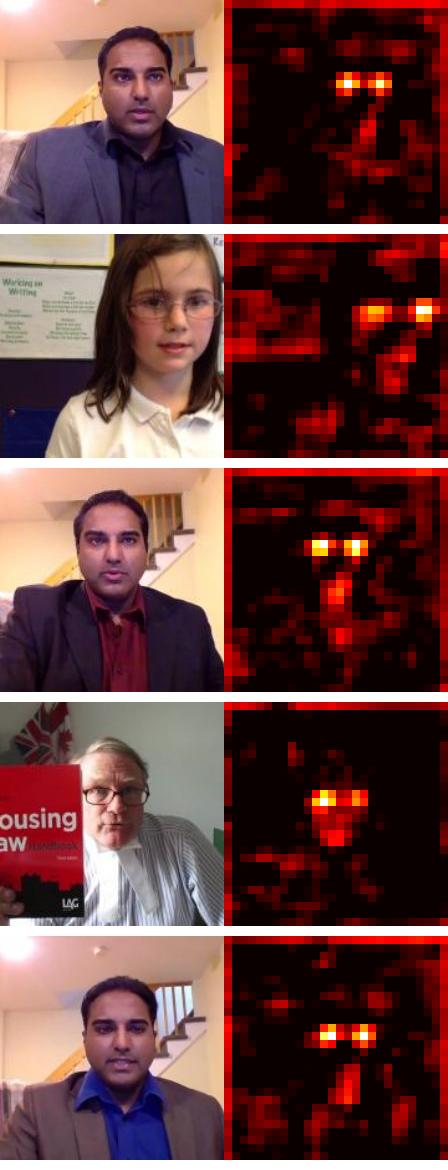} &
  \includegraphics[width=\unitsW]{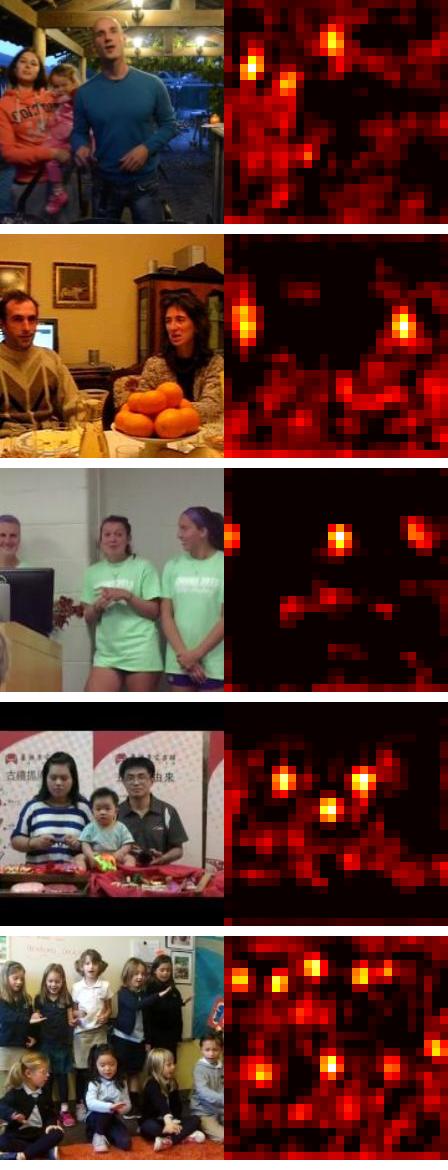} &
  \includegraphics[width=\unitsW]{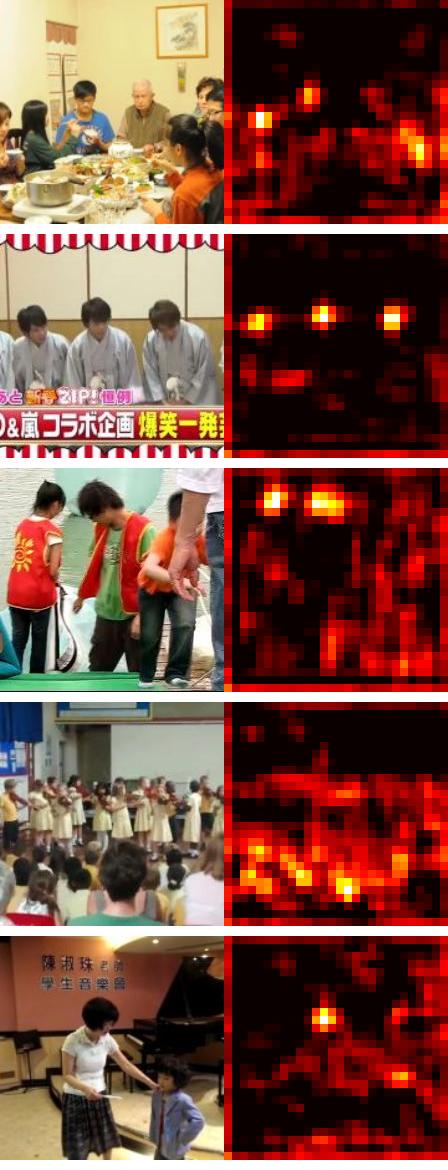} &
  \includegraphics[width=\unitsW]{figures/qual/soundnet500k_233004/vision_009.jpg}
  \\
\end{tabular}
\end{center}
   \caption{{\bf Learnt human-related visual concepts and semantic heatmaps (Flickr-SoundNet).}
Each mini-column shows five images that most activate a particular unit of the 512 in
\texttt{pool4} of the vision subnetwork, and the corresponding heatmap
(for more details see Figures \ref{fig:unitsV} and \ref{fig:unitsVmap}).
Column titles are a subjective names of concepts the units respond to.
}
\label{fig:unitsVSoundNetB}
\end{figure*}
}

\newcommand{\figUnitsASoundNetA}{
\def\unitsW{0.1\linewidth}
\def\fntsz{\scriptsize}
\begin{figure*}[p]
\begin{center}
\setlength{\tabcolsep}{2pt}
\begin{tabular}{cccccccc}
  \multicolumn{2}{c}{Baby voice} &
  \multicolumn{2}{c}{Human voice} &
  Male voice &
  People &
  \multicolumn{2}{c}{Crowd}
  \\
  \href{http://youtu.be/MrADE51ahHQ}{\includegraphics[width=\unitsW]{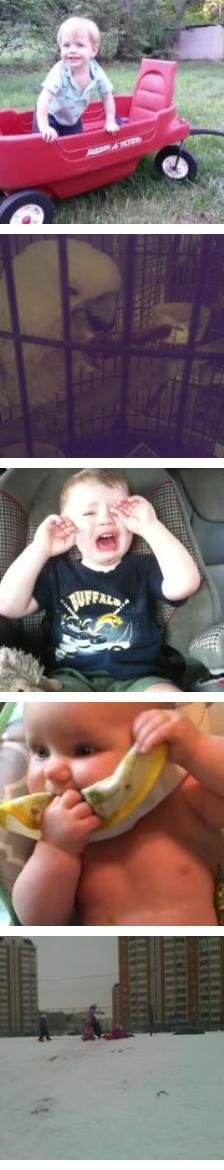}} &
  \href{http://youtu.be/L-SbBq2xJVo}{\includegraphics[width=\unitsW]{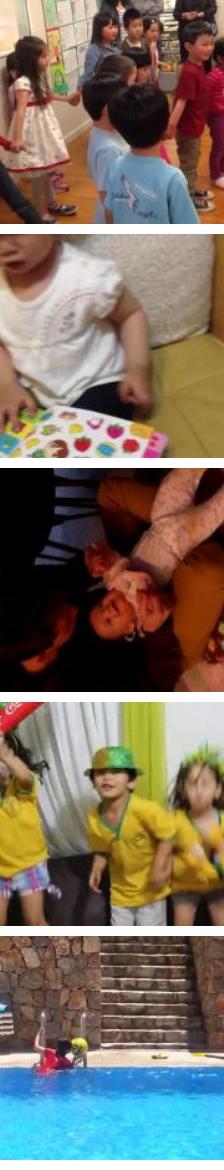}} &
  \href{http://youtu.be/i8khQ-MY9Qg}{\includegraphics[width=\unitsW]{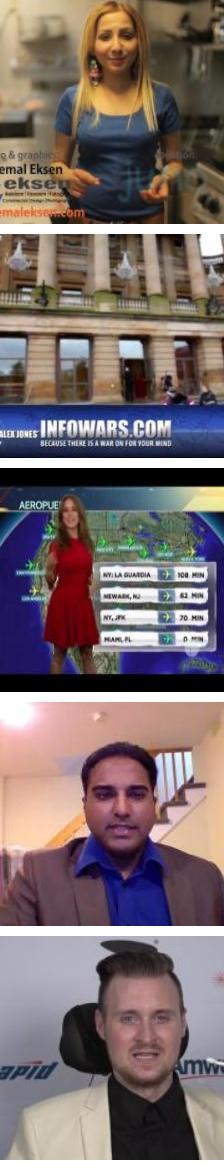}} &
  \href{http://youtu.be/ET3i_yZm4Ww}{\includegraphics[width=\unitsW]{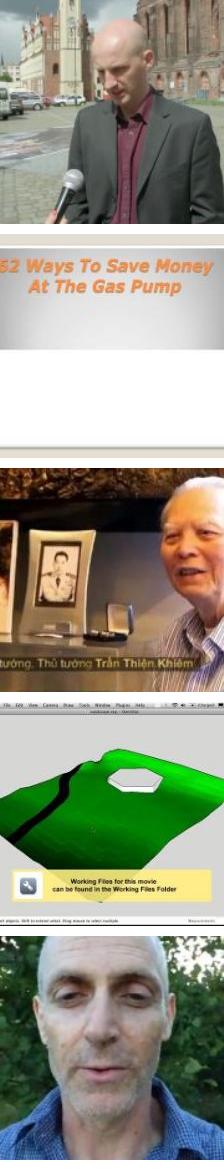}} &
  \href{http://youtu.be/vs5-5u2C-KI}{\includegraphics[width=\unitsW]{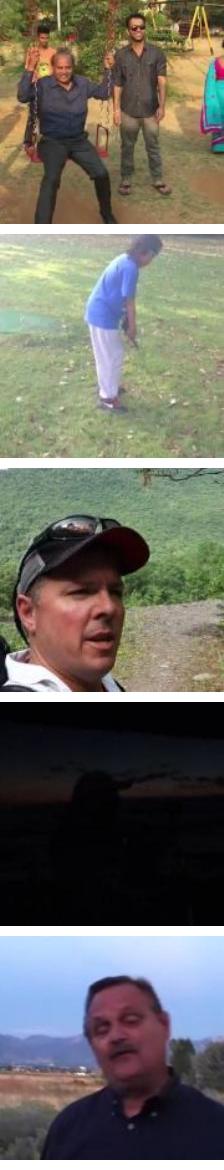}} &
  \href{http://youtu.be/GBL0475ctoQ}{\includegraphics[width=\unitsW]{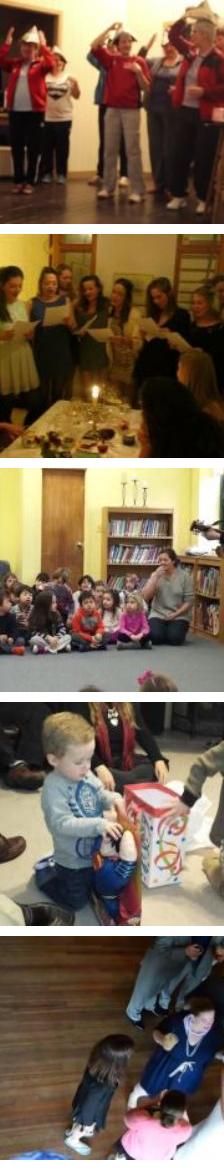}} &
  \href{http://youtu.be/Ax6Xcqn1Gxc}{\includegraphics[width=\unitsW]{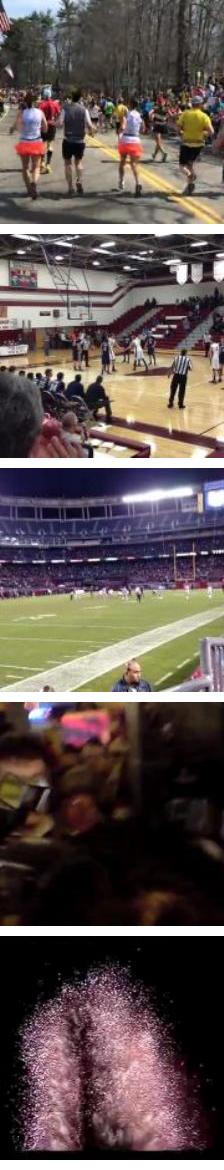}} &
  \href{http://youtu.be/Tt41s6xezkM}{\includegraphics[width=\unitsW]{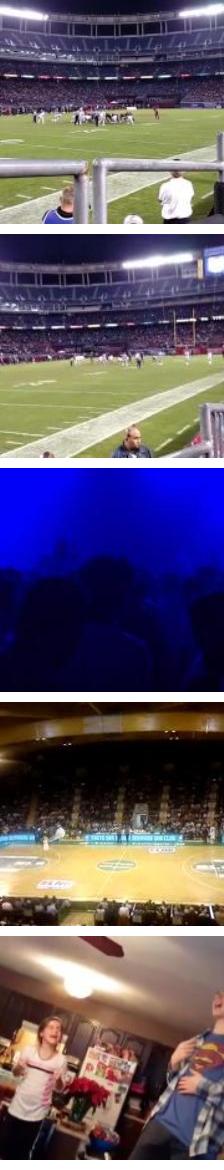}}
  \\ \hline \\
  Music &
  Concert &
  Sport &
  Clapping &
  Water &
  Underwater &
  Windy &
  Outdoor
  \\
  \href{http://youtu.be/A5aEXSUbGjg}{\includegraphics[width=\unitsW]{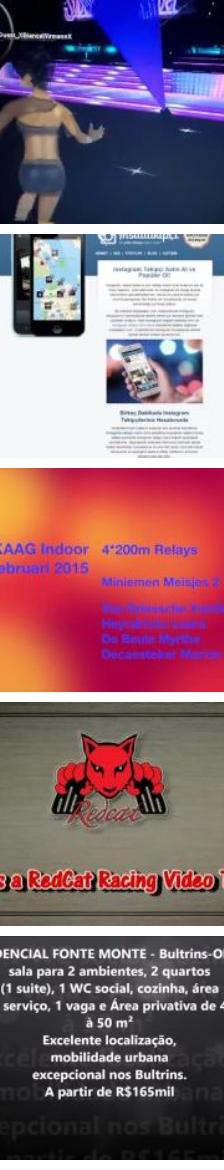}} &
  \href{http://youtu.be/cqQY7TrqQdw}{\includegraphics[width=\unitsW]{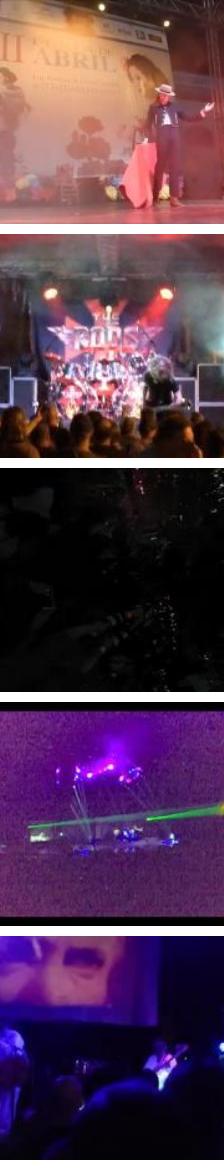}} &
  \href{http://youtu.be/8BPQd_DytyI}{\includegraphics[width=\unitsW]{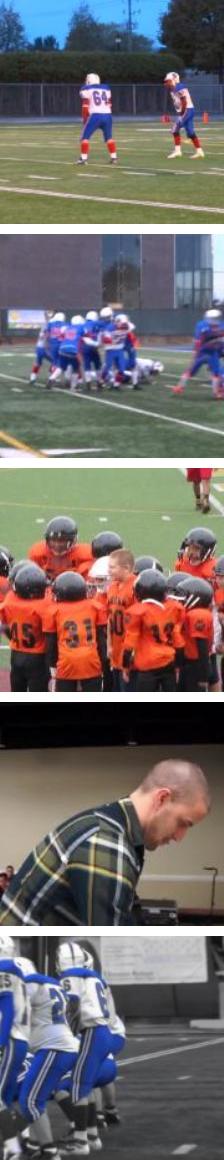}} &
  \href{http://youtu.be/GlbhKNqNCGw}{\includegraphics[width=\unitsW]{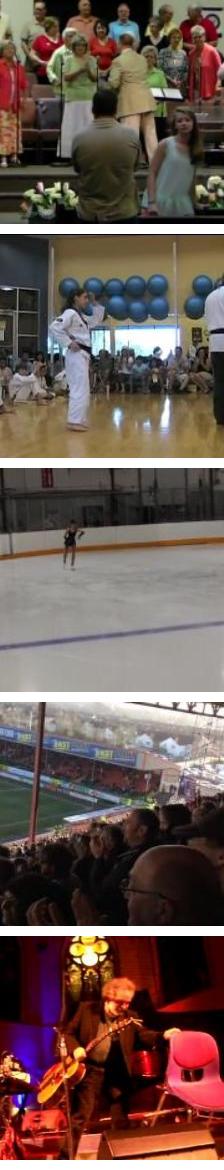}} &
  \href{http://youtu.be/GWW5scO468o}{\includegraphics[width=\unitsW]{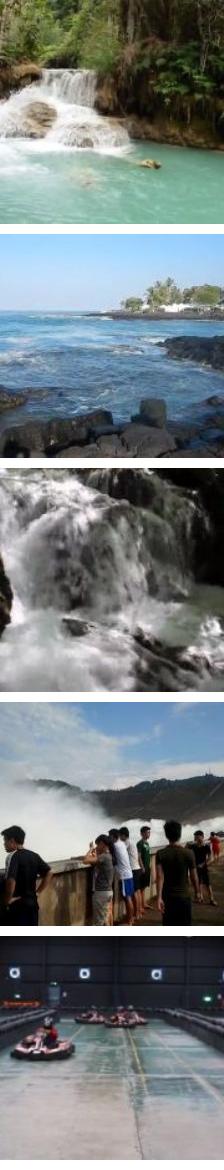}} &
  \href{http://youtu.be/_AR2dCWd3aY}{\includegraphics[width=\unitsW]{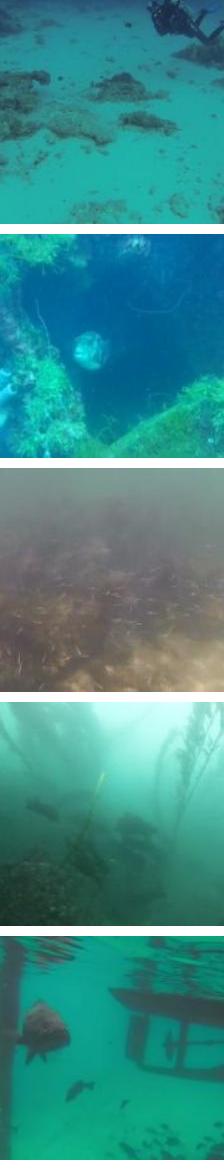}} &
  \href{http://youtu.be/gyIdY6ncxpc}{\includegraphics[width=\unitsW]{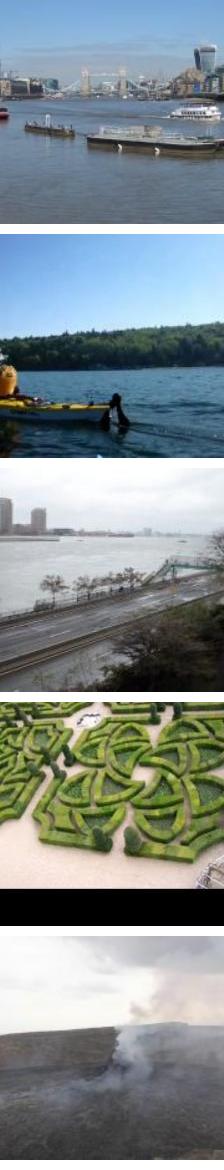}} &
  \href{http://youtu.be/7DxHhEewhwA}{\includegraphics[width=\unitsW]{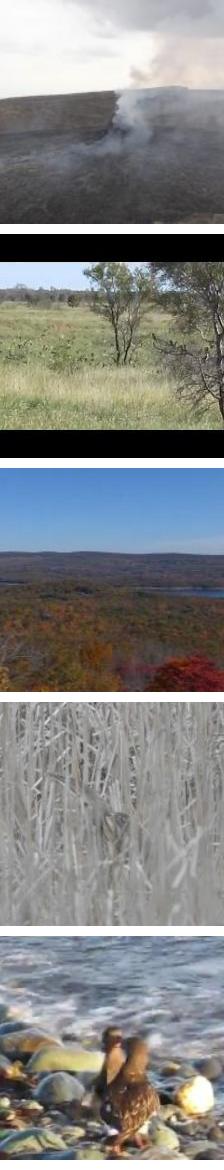}}
\end{tabular}
\end{center}
   \caption{{\bf Learnt audio concepts (Flickr-SoundNet).}
Each mini-column shows sounds that most activate a particular unit of the 512 in
\texttt{pool4} of the audio subnetwork.
Purely for visualization purposes, as it is hard to display sound,
the frame of the video that is aligned with the sound is shown instead
of the actual sound form,
but we stress that no vision is used in this experiment.
Column titles are a subjective names of concepts the units respond to.
Note that for the ``Human voice'', ``Male voice'', ``Crowd'', ``Music''
and ``Concert'' examples, the respective clips do contain the relevant audio
despite the frame looking as if it is unrelated, \eg the third example in the
``Concert'' column does contain loud music sounds.
Audio clips containing the five concatenated 1s samples corresponding to each
mini-column are hosted on YouTube and can be reached by clicking on the
respective mini-columns;
\href{https://goo.gl/ohDGtJ}{this YouTube playlist (https://goo.gl/ohDGtJ)}
contains all 16 examples.
}
\label{fig:unitsASoundNetA}
\end{figure*}
}

\newcommand{\figTsne}{
\begin{figure*}[p]
\begin{center}
\setlength{\tabcolsep}{2pt}
\hspace*{-0.5cm}
\begin{tabular}{c@{\hskip 1cm}cc}
  (a) Audio Learnt &
  \multirow{4}{*}{
\rule{0pt}{12cm}
  \includegraphics[trim={1cm 2cm 15cm 2.5cm},clip,height=0.4\linewidth]{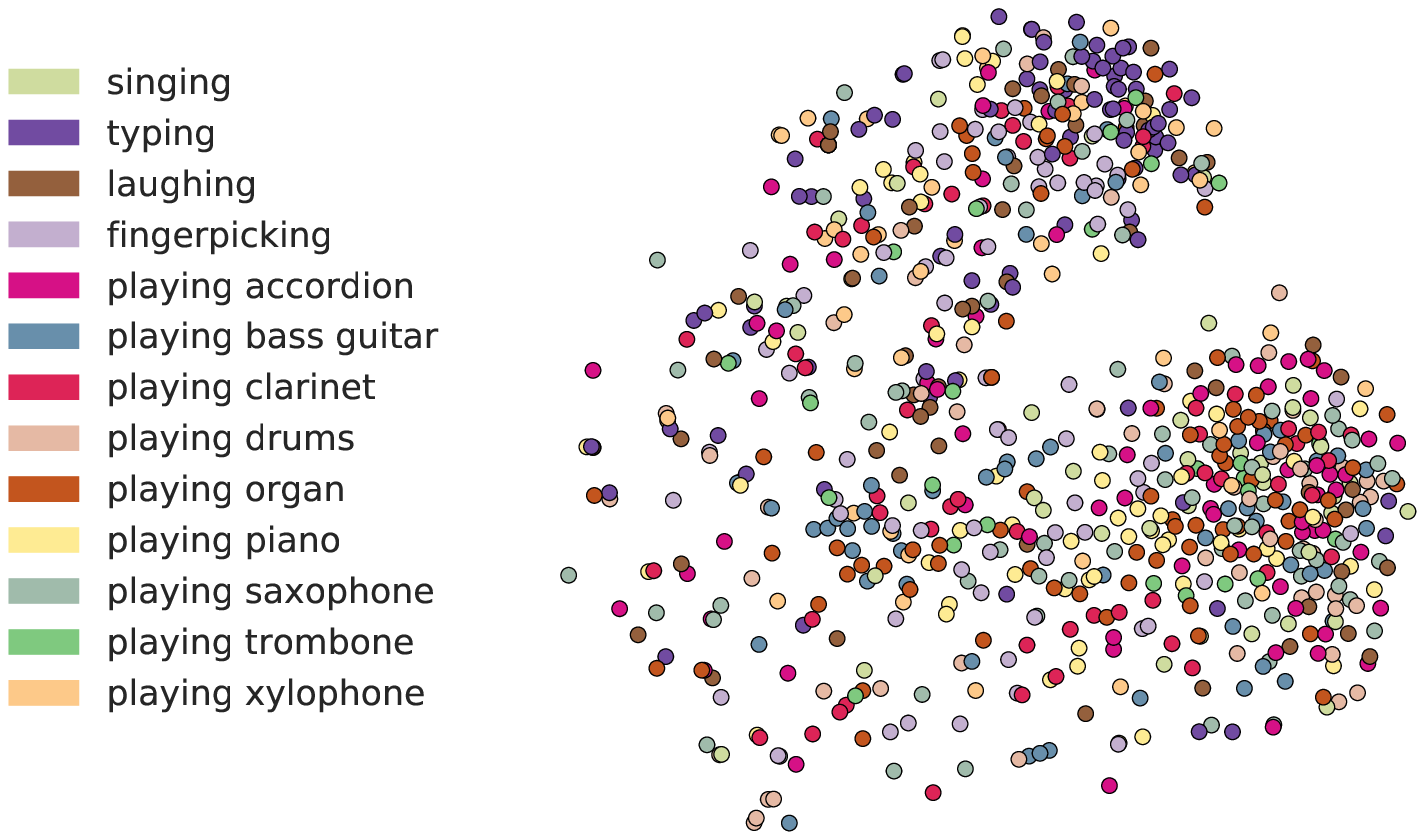}
  } &
  (b) Audio Random \\ [1em]
  \includegraphics[height=0.4\linewidth]{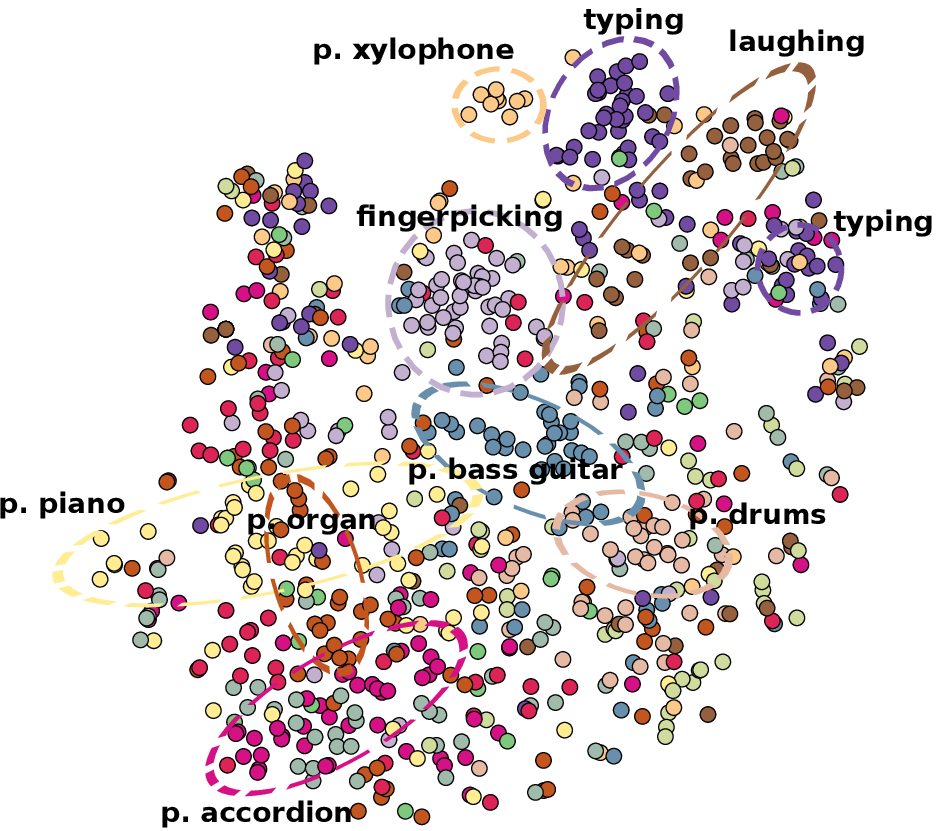} & &
  \hspace{-0.3cm}\includegraphics[trim={6.5cm 2.7cm 5cm 2.7cm},clip,height=0.35\linewidth]{figures/tsne/audio_15_random.eps} \\ [5em]
  (c) Visual Learnt &&
  (d) Visual Random \\ [1em]
\includegraphics[height=0.4\linewidth]{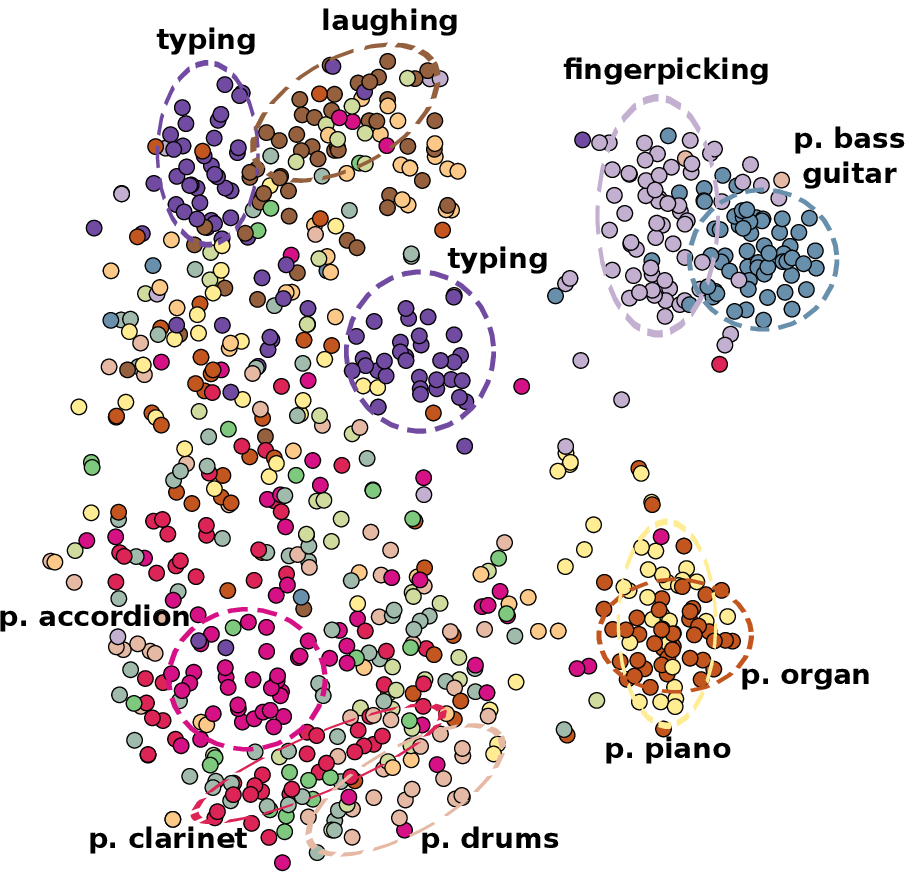} & &
  \hspace{-2.5cm}\includegraphics[trim={3cm 2cm 3cm 2cm},clip,height=0.33\linewidth]{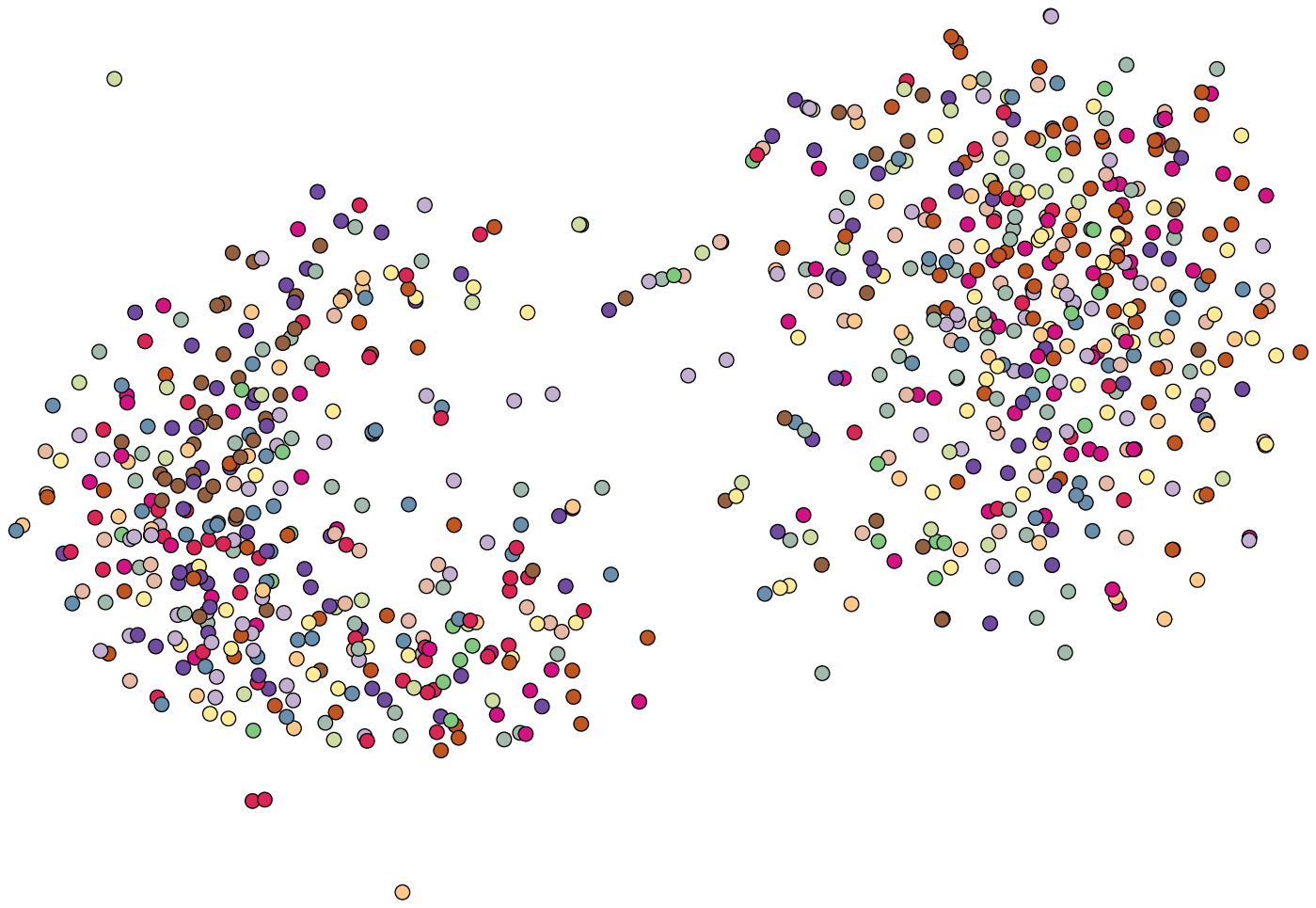}
\end{tabular}
\end{center}
   \caption{{\bf t-SNE visualization \cite{Van-der-Maaten08} of learnt representations (Kinetics-Sounds).}
The (a,c) and (b,d) show the two-dimensional t-SNE embeddings for the trained
versus non-trained (\ie network with random weights) $L^3$-Net, respectively.
For visualization purposes only, we colour the t-SNE embeddings using the
Kinetics-Sounds labels, but no labels were used for training the $L^3$-Net.
For clarity and reduced clutter, only a subset of actions (13 classes out of 34)
is shown. Some clearly noticeable clusters are manually highlighted by
enclosing them with ellipses.
Best viewed in colour.
}
\label{fig:tsne}
\end{figure*}
}

\newcommand{\tabNMI}{
\begin{table}[b!]
\begin{center}
\begin{tabular}{lcc}
  Method & Vision & Audio \\
  \hline\hline
  Random assignments & 0.165 & 0.165 \\
  Ours random ($L^3$-Net without training) & 0.204 & 0.219 \\
  Ours ($L^3$-Net self-supervised training) & 0.409 & 0.330
\end{tabular}
\end{center}
    \caption{{\bf Clustering quality.}
Normalized Mutual Information (NMI) score between the unsupervised clusterings of
feature embeddings and the Kinetics-Sounds labels.
}
\label{tab:nmi}
\end{table}
}